\documentclass[nohyperref]{article}

\usepackage{microtype}
\usepackage{graphicx}
\usepackage{subfigure}
\usepackage{booktabs} % for professional tables

\usepackage{hyperref}

\usepackage[accepted]{arxiv}

\usepackage{amsmath}
\usepackage{amssymb}
\usepackage{mathtools}
\usepackage{amsthm}
\newcommand{\exam}[2]{{\tt #1} $\Rightarrow$ {\tt #2}}

\usepackage{graphicx}
\usepackage{float}

\definecolor{mygray}{gray}{0.5}
\definecolor{cblue}{RGB}{8, 85, 153}
\definecolor{darkblue}{RGB}{1, 43, 112}
\newcommand{\cblue}[1]{{\textcolor{cblue}{#1}}}
\definecolor{cgreen}{RGB}{8, 153, 83}

\usepackage{amsmath}
\usepackage{amssymb}
\usepackage{amsthm}
\usepackage{booktabs}
\usepackage{longtable}
\usepackage{placeins} %FloatBarrier
\usepackage{makecell}
\usepackage{multirow}

\usepackage[capitalize]{cleveref}
\usepackage{fontawesome}

\usepackage{adjustbox}

\usepackage{tabularx}

\newcommand{\method}{iPrompt}
\newcommand{\methodit}{\textit{\method}}
\newcommand{\methods}{iPrompt }
\newcommand{\methodlongs}{interpretable autoprompting }

\newcommand{\blank}{\underline{\hspace{15pt}}}

\theoremstyle{definition}

\crefname{definition}{Definition}{Definitions}%
\crefname{section}{Sec.}{Secs.}%
\usepackage{scrextend} % for addmargin
\theoremstyle{plain}

\theoremstyle{definition}

\theoremstyle{remark}

\usepackage[textsize=tiny]{todonotes}

\icmltitlerunning{iPrompt\hfill\thepage}

\begin{document}

\twocolumn[

% \title{Understanding data via interpretable autoprompting of large language models}
% \title{Towards Scientific Discovery with Language Models via interpretable autoprompting}
% \title{Explaining Patterns in Data  with  Language Models via Interpretable Autoprompting} % Uncovering
\icmltitle{iPrompt: Explaining Data Patterns in Natural Language\\via Interpretable Autoprompting} % Uncovering
% \icmltitle{Submission and Formatting Instructions for \\
           % International Conference on Machine Learning (ICML 2022)}



\icmlsetsymbol{equal}{*}

\begin{icmlauthorlist}
\icmlauthor{Chandan Singh}{equal,a}
\icmlauthor{John X. Morris}{equal,b}
\icmlauthor{Jyoti Aneja}{a}
\icmlauthor{Alexander M. Rush}{b}
\icmlauthor{Jianfeng Gao}{a}

\end{icmlauthorlist}

\icmlaffiliation{a}{Microsoft Research}
\icmlaffiliation{b}{Cornell University}
\icmlcorrespondingauthor{Jianfeng Gao}{jfgao@microsft.com}
\icmlkeywords{Machine Learning, ICML}

\vskip 0.3in
]



\printAffiliationsAndNotice{\icmlEqualContribution} % otherwise use the standard text.

\begin{abstract}
Large language models (LLMs) have displayed an impressive ability to harness natural language to perform complex tasks. We explore whether we can leverage this ability to find and explain patterns in data. Specifically, given a pre-trained LLM and data examples, we introduce interpretable autoprompting (\methodit), an algorithm that generates a natural language string explaining the data. iPrompt iteratively generates explanations with an LLM and reranks them based on their performance when used as a prompt.
Experiments on a wide range of datasets, from synthetic mathematics to natural language understanding, show that iPrompt can yield meaningful insights by accurately finding dataset explanations that are human-interpretable.
% Moreover, \methods is reasonably efficient, as it does not require access to model gradients and works with relatively small models (e.g. ~6 billion parameters rather $\geq$100 billion).
On two of four classification datasets, \methods discovers a prompt that outperforms human-written prompts on GPT-3,
despite only querying the relatively small GPT-J model.
Finally, experiments with scientific datasets show the potential for iPrompt to aid in scientific discovery.
\footnote{All code for using the methods and data here is made available on Github.
% at \href{https://github.com/csinva/interpretable-autoprompting}{\faGithub\,github.com/csinva/iprompt} and a simple API is available in the imodelsX package: \href{https://github.com/csinva/imodelsX}{\faGithub\,github.com/csinva/imodelsX}.
% We demonstrate using these explanations for various tasks, such as describing an unknown function of finding differences between datasets.
% Moreover, the prompts produced by iPrompt are simultaneously human-interpretable and highly effective for generalization: on real-world sentiment classification datasets, iPrompt produces prompts that match or even improve upon human-written prompts for GPT-3.
}
\end{abstract}

\section{Introduction}

Large language models (LLMs) have attained an extraordinary ability to harness natural language for solving diverse problems~\citep{devlin2018bert},
often without the need for finetuning~\citep{brown2020language,sanh2021multitask}.
Moreover, LLMs have demonstrated the capacity to excel at real-world problems, such as mathematics~\citep{lewkowycz2022solving}, scientific question answering~\citep{sadat2022scinli},  general processing of scientific text~\citep{beltagy2019scibert}, predicting brain responses~\cite{schrimpf2021neural}, and classifying proteins and chemical compounds~\citep{taylor2022galactica}.
% (and related neural-network models) have demonstrated the capacity to excel at real-world problems, such as mathematics~\citep{lewkowycz2022solving}, controller design for nuclear fusion~\citep{degrave2022magnetic}, and even at tasks beyond the ability of humans, such as protein-folding~\citep{jumper2021highly} and neural data representation~\citep{jain2018incorporating}.}

In this work, we probe whether we can leverage the learned skills of an LLM to \textit{discover and explain patterns} in a dataset.
To do so, we invert the typical problem of fitting an LLM to data
and instead ask whether we can use a fixed LLM to produce a natural language string explaining dataset patterns.
% describing the underlying process in a dataset.
% This problem formulation enables the possibility of potentially generating new scientific hypotheses, particularly in cases where an LLM can identify structure in data that eludes humans.

Our approach to this problem centers around prompting.
Prompting has emerged as an effective method for adapting LLMs to new datasets~\citep{liu2021pre};
a prompt string is combined with each example in a dataset before querying an LLM for an answer.
While prompts were initially constructed manually, recent work has shown success in \textit{autoprompting}, automatically finding a prompt via optimization~\citep{shin2020autoprompt,li2021prefix,deng2022rlprompt}.
However, previous work on learning natural language prompts does not produce prompts that are meaningful to humans.
% Moreover, previous work focuses on finding prompts used to describe a task, rather than a dataset.

\begin{figure}[t]
    \centering
    % \vspace{-12pt}
    \includegraphics[width=0.8\columnwidth]{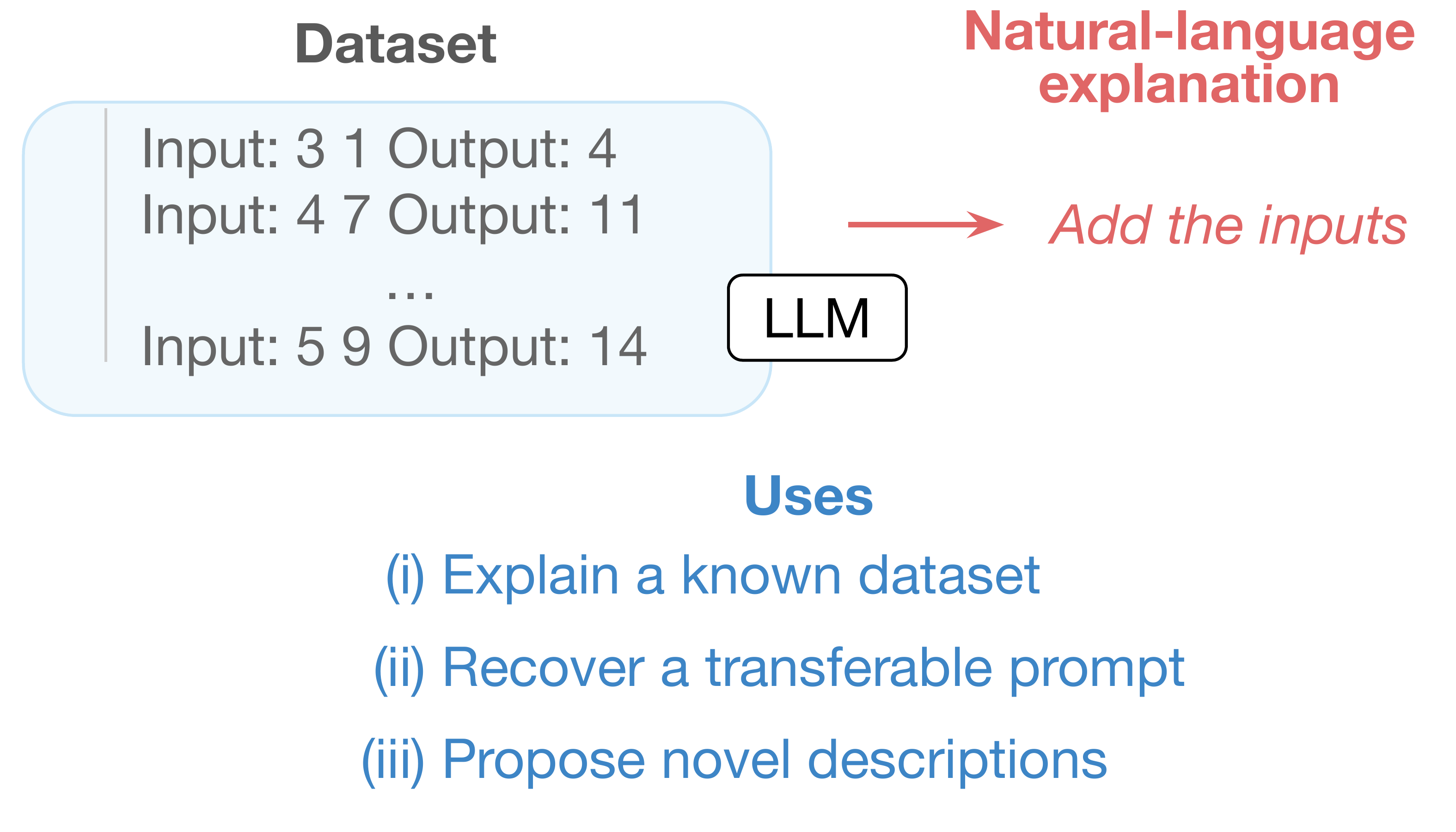}
    \vspace{-5pt}
    \caption{Interpretable autoprompting (\methodit) inverts the standard prediction problem to instead find a natural language explanation of the data using a fixed, pre-trained large language model.
    }
    \label{fig:intro}
\end{figure}

Our approach, \methodlongs (\methodit), extends autoprompting to generate a semantically meaningful natural language prompt that explains a key characteristic of the data (see \cref{fig:intro}).
For example, given a dataset of examples of addition, e.g. \exam{2 5}{7} ... \exam{3 1}{4}, \methods yields the natural language explanation \textit{Add the inputs}.
By changing the input form of the data, we can generate explanations that accomplish different tasks from the example, such as: i) recovering a dataset explanation, ii) generating a prompt transferable between LLMs, and iii) proposing novel descriptions.
\methods works by using a pre-trained LLM to iteratively propose and evaluate different candidate explanations.
% without requiring access to the LLM's gradients.
% \methods is an iterative algorithm that alternates between (i) proposing candidate explanations with an LLM, (ii) reranking the candidates based on their performance when used as a prompt, and (iii) exploring new candidates.
% Through the use of natural language, \methods provides flexible explanations of diverse data and can incorporate prior human knowledge directly, by adding it into the prompt.

For evaluation, we curate a diverse collection of datasets written in natural language (\cref{tab:dataset_ovw}) and measure \method's ability to accurately explain a ground-truth pattern.
% The dataset includes a number of synthetic math datasets,
% as well as language tasks from
% the Natural Instructions V2 dataset~\citep{naturalinstructionsv2}.
We find that \methods 
outperforms baseline methods in accurately finding a correct description;
% across these datasets.
moreover, the generated descriptions are interpretable, allowing human auditing and enabling strong generalization when used as a prompt in a new setting (i.e. when used for a different LLM). 
On real-world sentiment classification datasets,
\methods even produces prompts that match or improve upon human-written prompts for GPT-3, while only using smaller, locally-run language models.
Finally, we find that \methods is able to extract information from real-world scientific datasets. % (\cref{sec:science}).
% , particularly a neuroscience dataset
% in which we seek to understand the mapping of semantic concepts in the brain from fMRI imaging (\cref{sec:science}).

% In what follows, \cref{sec:background} covers the background and related work, \cref{sec:methods} describes the algorithms we use for interpretable autoprompting, \cref{sec:results} shows the results, and \cref{sec:discussion} concludes with a discussion.

\section{Related work}
\label{sec:background}

% \cs{TODO: cite "fantastically ordered prompts" and "describving difference between two distributions"}
\paragraph{Prompting and autoprompting.}
With the advent of large-scale models, prompting (i.e. finding the right prompt to use to query an LLM for a given task) has exploded as an area of inquiry, often yielding impressive improvements in performance~\citep{brown2020language,petroni2019language,liu2021pre} and spurring a line of work aiming to make prompting easier~\citep{strobelt2022interactive,lu-etal-2022-fantastically,bach2022promptsource,logan-iv-etal-2022-cutting}.
Recently, autoprompting (i.e. automatically searching for a prompt or prompt-embedding via optimization) has emerged,
% to improve the process of prompting,
with methods such as prefix-tuning~\citep{li2021prefix}, P-tuning~\citep{liu2021gpt}, prompt-tuning with rules~\citep{han2021ptr}, knowledgeable prompt tuning~\citep{hu2021knowledgeable} and many more~\citep{liu2021pre}.
These strategies use gradient descent to find a set of ``adapter'' parameters that maximize model performance, but do not require that the new parameters map back to tokens in discrete space, rendering them uninterpretable.
% \paragraph{Prompting in discrete space.}

A few methods tackle the more difficult problem of searching for prompts that can be expressed in natural language tokens.
RLPrompt~\citep{deng2022rlprompt} searches for such a prompt using reinforcement learning and one recent work~\citep{honovich2022instruction} queries an LLM to produce a prompt.
AutoPrompt~\citep{shin2020autoprompt} performs autoprompting via input gradients (see \cref{sec:methods}).
Similarly, adversarial triggers~\citep{wallace2019universal} use autoprompting to identify adversarial inputs which can be used to change a model's prediction.
These methods effectively alter a model's predictions,
but do not constrain the discovered prompts to be semantically meaningful,
resulting in prompts that are difficult to interpret~\citep{webson2021prompt}.
Another related work directly finetunes an LLM to describe the difference between two datasets~\citep{zhong2022describing}.
% We share common methodology with these works, but reformulate the problem and approach to ensure the generated prompt is human-interpretable.
Concurrent work proposes a method for natural language prompting similar to the one here, with a focus on improving prediction performance rather than on explaining data patterns~\cite{zhou2022large}.

\paragraph{Problems related to dataset explanation}
The problem statement presented in this work closely resembles the widely studied problems of symbolic regression~\citep{augusto2000symbolic,schmidt2009distilling}, program synthesis~\citep{gulwani2017program,manna1980deductive}, text/table  summarization~\citep{kryscinski2019neural,liu2018table}, and pattern discovery in data-mining~\citep{hand2007principles}.
% In these cases, data examples are given with the goal of inferring a symbolic expression, program, or text summary that is consistent with the data.
\methods can be viewed as an algorithm for symbolic regression, in which the set of allowable symbols consists of semantically meaningful natural language strings.
One recent work proposes the task of inferring prompts that improve supervised prediction~\citep{honovich2022instruction}, which we generalize here to diverse use cases for dataset explanation.
% and their optimization is guided by a pre-trained LLM.

\paragraph{Alternative methods for neural-network interpretation}
A popular method for interpreting neural networks is to inspect an LLM's individual predictions via
feature importances~\citep{lundberg2019explainable,ribeiro2016should},
feature-interaction importances~\citep{singh2019Hierarchical,tsang2017detecting},
extractive rationales~\citep{zaidan2008modeling,sha2021learning},
or natural language explanations for individual predictions~\citep{hendricks2016generating,camburu2018snli}.
These works can provide meaningful insights for individual predictions, but it is difficult to parse them into an understanding of an entire dataset.
Alternatively, one can investigate whether an LLM's learned representations via probing~\citep{conneau2018you,liu2019incorporating}
% (which tests whether certain properties can be decoded from LLMs embedding)
or by directly analyzing a model's internal weights and activations~\citep{wang2021inferbert,olah2018building,meng2022locating}.
However, these approaches are limited in their ability to generate previously unknown descriptions of data.
A different approach involves distilling information into a transparent model \cite{tan2018distill,ha2021adaptive,singh2022embgam} or simply using a transparent model in the first place~\cite{breiman1984classification,tan2022Fast,singh2021imodels,agarwal2022Hierarchical}.
% Even simpler, some works forego the black-box model altogether and introduce structure into a DNN to make it more interpretable, e.g. by using prototypes~\citep{li2018deep,chen2019looks} or identifying intermediate concepts in a network~\citep{koh2020concept}.

\section{Methods: Defining the task and approach}
\label{sec:methods}

\subsection{Task: Dataset Explanation}
\label{sec:task_datasets}
% \paragraph{Task definition} 
Given a dataset %$\mathcal{D}$
comprised of input-output string pairs $\{(x^1, y^1), \ldots, (x^N, y^N)\}$,
the goal is to produce a ``semantically meaningful'' natural language string that explains the relationship between
% the underlying process used to generate
$x$ and $y$.
% pattern in the data.
% $\mathcal{T}$.
% Each example $X_i$ contains at least a single demonstration an input-output relationship (e.g. ``In: 3, 5 Out: 8'').
% Notably, no prompt or task definition is provided, and the goal is to uncover 
We require that a string consists of human-understandable text rather than a sequence of incongruous tokens.
For example, in the dataset shown in \cref{fig:intro}, given samples of data performing addition, our task is to recover text synonymous to \textit{Add the inputs}.
This dataset explanation can then be used for various downstream tasks, such as prompting a different LLM.
% We study two settings of the task: in one we are given a template string and in the other we are not.
% We also consider a more constrained setting where a user-specified \textit{template} string is provided.
% For example a generic template may be \textit{To get the output from the inputs, \blank}.
% Generic templates allow a model to infer missing words from a broad set of relationships. In both settings, methods  generate a ranked set of candidate explanations. 
% Srush : I didn't understand this distinction
% whereas specific templates allow a user to leverage prior knowledge to elicit more specific details e.g. \textit{To compute the output, the most important input is \blank}.
% This task is a generalization of that proposed by \citeauthor{honovich2022instruction}, which they call ``instruction induction''.

% Dataset explanation can be used for more than prompting LLMs.
% it can be used to describe the difference between two datasets for which the 
% For example, by drawing $x^\textit{(i)}$ from different datasets and setting $y^\textit{(i)}$ as \textit{Yes} or \textit{No} depending on which dataset it came from, the dataset explanation can be used to describe the difference between two datasets (e.g. distribution shift, function change, change in an underlying property).

\paragraph{Datasets}
\label{subsec:datasets}
\cref{tab:dataset_ovw} shows the collections of datasets we study:
% The collections consist of three different types of datasets:
(1) \textit{Synthetic math} -- datasets that require inferring an underlying mathematical function based on numeric input and outputs;
(2) \textit{Allen NLI} (ANLI) and (3) \textit{Instruction induction}~\citep{honovich2022instruction} -- diverse language tasks~\citep{naturalinstructionsv2} with easily verifiable descriptions (e.g. \textit{Find a country's capital}). 
 % -- language tasks with groundtruth descriptions,
% constructed with the goal of inferring easily verifiable prompts,
% (4) \textit{Distribution differences}, text binary classification with known differences~\citep{zhong2022describing,zhong2021adapting},
(4) \textit{Sentiment} -- a collection of sentiment classification datasets in different domains.
% (4) fMRI, a dataset involving brain responses to natural language~\citep{huth2016natural}, motivated by the goal of recovering unknown explanations. 
For collections (1-3), there is a ground-truth prompt available for evaluation. For example, when adding two numbers (\cref{fig:intro}), the rule checks whether a description contains any of the keywords \textit{add}, \textit{sum}, or \textit{+}.
We also study scientific datasets on (5) \textit{proteins/chemicals}, and (6) \textit{fMRI} with full details given in \cref{sec:science}.

\begin{table}[t]
    \centering
    \footnotesize
    \caption{Dataset Explanation Tasks.
    Each collections contains \# different task. 
    Roman numerals correspond to the use cases in \cref{fig:intro}.
    For full details on each dataset, see \cref{subsec:data_datails_supp}.}
    
    % \makebox[\textwidth][c] {
    % \begin{tabular}{cccc}
%     \toprule
%      Collection & \# & Description & Dataset names\\
%     \midrule 
%      \vspace{2pt}
%      \makecell[t]{Inverse synthetic\\math}& 10 & \makecell[t]{Simple mathematical\\functions} & \makecell[t]{Add two, Subtract two, Multiply two,\\
%      Divide two, Max two, First number,\\Square, Exponentiate, Double, Fibonacci}\\
%      \vspace{2pt}
%      \makecell[t]{Inverse Allen NLI\\\citep{naturalinstructionsv2}} & 10 & Diverse language tasks & \makecell[t]{Country capital, Antonyms, Check edibility,\\Rhyme generation, Country currency,\\Check prime, Check vegetarian, Find typo,\\Gender classification, SQL query generation}\\
%      \vspace{2pt}
     
%      \makecell[t]{Sentiment} & 4 &\makecell[t]{Sentiment classification} & \makecell[t]{SST-2, RottenTomatoes, IMDB, \\ Financial Phrasebank}\\ %Tweet (Hate),
%      \vspace{2pt}
     
%      \makecell[t]{Natural-language fMRI\\\citep{huth2016natural}} & 20 &\makecell[t]{Categorize a list of words\\that excite an fMRI voxel} & \makecell[t]{Extracting a pattern from a set of\\ words, each corresponding\\to a different voxel}\\
%     %  (Inverse) Big-Bench & 20 & Simple NLI tasks. & \makecell{Logical deduction, ...}\\
%      \bottomrule
% \end{tabular}
\begin{tabular}{lcl}
    \toprule
     Collection & \# & Description \\ % & Dataset names\\
    \midrule 
     % \vspace{2pt}
     1) Synthetic math & 10 & Mathematical functions \cblue{(i), (ii)}\\% & \makecell[t]{Add two, Subtract two, Multiply two,\\
     % Divide two, Max two, First number,\\Square, Exponentiate, Double, Fibonacci}\\
     % \vspace{2pt}
     2) Allen NLI & 10 & Language tasks \cblue{(i), (ii)} \\ %& \makecell[t]{Country capital, Antonyms, Check edibility,\\Rhyme generation, Country currency,\\Check prime, Check vegetarian, Find typo,\\Gender classification, SQL query generation}\\
     3) Instr. induction & 20 & Language tasks \cblue{(i), (ii)} \\
     % \makecell[t]{Distribution differences} & 54 & Binary classification tasks \cblue{(ii)} \\
     % \vspace{2pt}
     % \vspace{2pt}
    4) Sentiment & 4 & Sentiment classification \cblue{(i), (ii)}\\ %& \makecell[t]{SST-2, RottenTomatoes, IMDB, \\ Financial Phrasebank}\\ %Tweet (Hate),
     \midrule
     5) Proteins/chemicals & 3 & Protein/chemical properties \cblue{(iii)} \\     
     6) Language fMRI & 20 & Excitation of fMRI voxel \cblue{(iii),(iii)} \\
     \bottomrule
\end{tabular}%}
    \label{tab:dataset_ovw}
\end{table}

% For classification tasks such as \textit{Check edibility}, the label provided in the example text is simply \textit{Yes}/\textit{No} rather than the given labels, e.g. \textit{edible}/\textit{non-edible}.

\subsection{Approach: \methods}
We now detail approaches for the general problem of autoprompting before introducing \method, our method for interpretable autoprompting.
% in \cref{subsec:iprompt_method}.
We specify autoprompting as a discrete search problem. Given a dataset of $n$ input-output pairs \{($x^1$, $y^1$), ..., ($x^n$, $y^n$)\} and a pre-trained LLM $f$ that returns the log-probability of a given string, autoprompting finds a natural language explanation $\hat{s}$ maximizing:
\begin{align}
    \hat s = \underset{s \in \mathcal S}{\text{argmax }} \sum_{i=1}^n f\left( \text{render}(s, x^i, y^i) \right)
    \label{eq:objective}
\end{align}
% \end{definition}
The \text{render} function is a problem-specific function that renders a natural language string from the prompt $s$ and each example in the dataset $(x^i, y^i)$. We use $\mathcal S$ to indicate the set of fluent strings, under some notion of syntactic fluency. This constraint is used to ensure prompts are readable, and potentially generalize to downstream LLMs. 
Solving this search problem exactly is intractable.

\begin{figure}[t]
    \centering
    \includegraphics[width=\columnwidth]{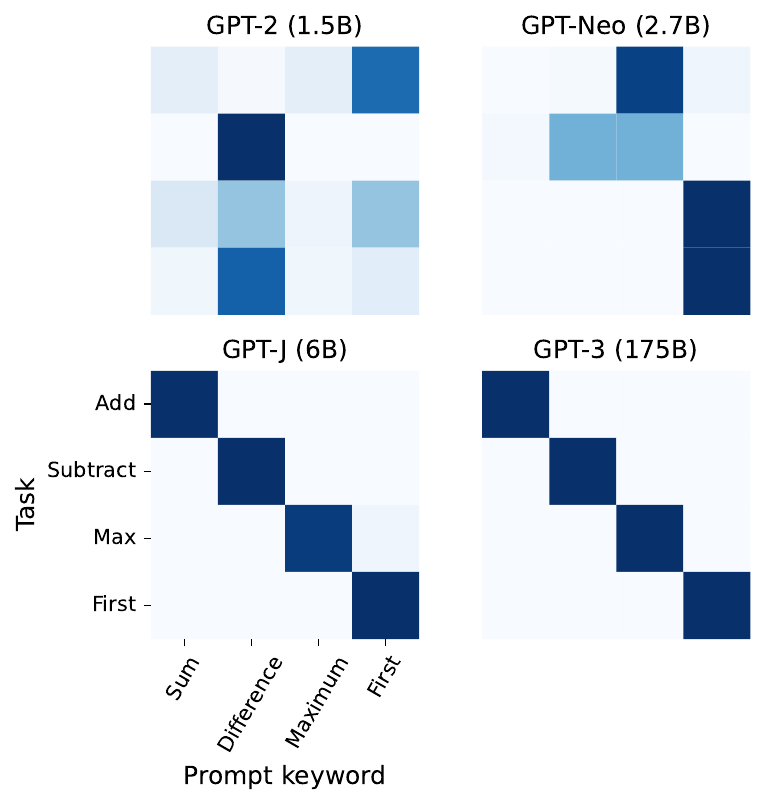}
    \vspace{-25pt}
    \caption{Prompt-based reranking depends on model size. 
    Large models (GPT-J 6B and GPT-3) align prompts correctly to tasks.
    The model is given the prompt \textit{Return the \blank of the inputs.}, where \blank \: is filled in with the shown prompt keyword before querying the output given two inputs numbers in a string.
    Darker indicates a higher accuracy, and high accuracy along the diagonal indicates that the correct prompt induces the highest accuracy.
    }
    \label{fig:prompt_selection}
\end{figure}

A core assumption of this objective is that semantically accurate prompts lead a model to assign higher probability to the correct output.
To check this assumption, we analyze four datasets from the inverse synthetic math collection that share common structure for the inputs and prompts. Each dataset admits a prompt of the form \textit{Return the \blank\ of the inputs.}, then is given two input numbers and queried for the output.

\cref{fig:prompt_selection} shows the accuracy of different models at performing these tasks across different input prompts.\footnote{The accuracy is normalized for each task using softmax in order to visualize the effect of differing prompts.}
For small models, the prompts are unsuccessful, but for large models (GPT-J 6B and GPT-3),
the model is accurate if and only if given the correct prompt.\footnote{For details on each model, see \cref{tab:models}.}
This result suggests that, at least for large models, the search for a prompt that maximizes performance correlates well with the underlying task. We will see in \cref{fig:ablation_acc_vs_mrr} that dataset explanation depends on this ability. 

% \subsection{Baseline Methods}
% \label{subsec:method_algos}

\paragraph{Baseline: AutoPrompt}
AutoPrompt~\citep{shin2020autoprompt} targets the objective posed in \cref{eq:objective} using a gradient-based local search.
AutoPrompt searches for $\hat s$ following the gradients of the objective \cref{eq:objective} with respect to individual tokens in $\hat s$.
It discretely changes individual words in $\hat s$ and then checks whether or not the newly updated $\hat s$ improves the objective score.
The use of gradients allows AutoPrompt to find an effective prompt $\hat s$, but makes it difficult to find answers that satisfy the fluency constraint ${\cal S}$.

% \paragraph{Zero-Shot Prompt Generation} 
% \footnote{We denote this problem as zero-shot prompt generation (not one-shot) because the example the LLM receives (\textit{In: 2 5 Out: 7}) does not give it a demonstration for the data explanation task it is trying to perform.}
\paragraph{Baseline: Zero-shot suffix decoding}
LLMs themselves can be directly used to predict prompt strings.
Following \citeauthor{honovich2022instruction},
we give the model a prompt string which contains data examples (e.g. 
$\underbrace{\textit{In: 2 5}}_{x^i} \underbrace{\textit{Out: 7.}}_{y^i} \underbrace{\textit{To compute the output from the input, }}_{\textit{template}}$\blank,) and sample the output to recover a prompt $\hat s$ using nucleus sampling.\footnote{We also consider averaging the model's output logits across all examples in the dataset before decoding the output, but find that it does not improve performance~(see \cref{sec:exp_details_supp}).}
% Sampling directly from $f$ helps ensure that the generated explanation is fluent and semantically meaningful.
% We decode the output using beam search to find the highest-probability outputs for multi-token prompts.\footnote{Here we prefer beam search here over alternatives such as nucleus sampling~\citep{holtzman2019curious} as we are interested in finding an accurate prompt description with as few samples as possible.}
% To improve on this approach, we place several examples into the model's context, and then average the model's output logits across all the examples in the dataset before decoding the output, an approach we refer to as \textit{average-suffix decoding}.
% We find that this technique is insufficient to find high-scoring prompts.

\paragraph{Proposed method: \methods}
% \label{subsec:iprompt_method}

\begin{figure}[t]
    %\centering
    \vspace{-4pt}    \includegraphics[width=\columnwidth]{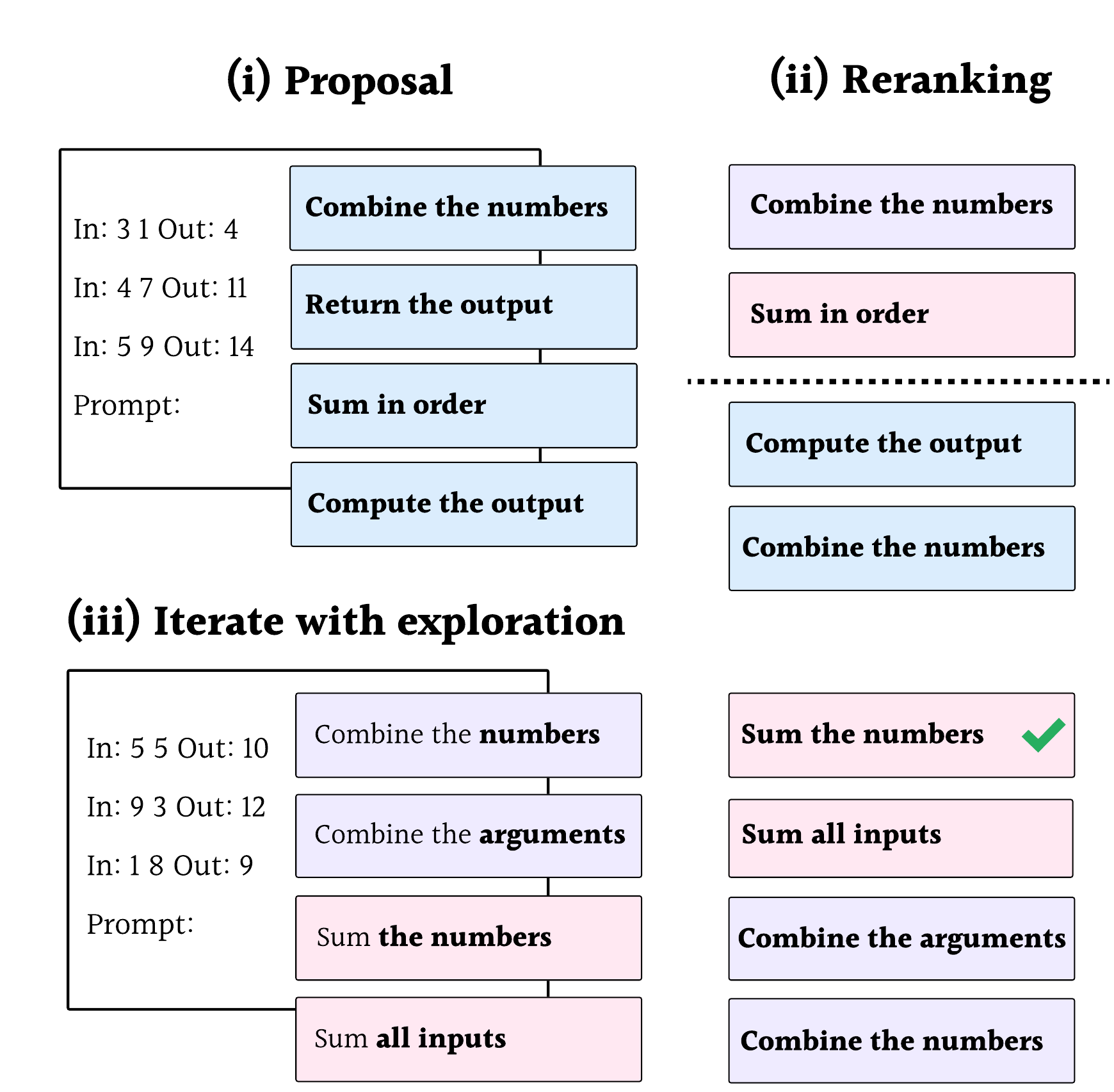}    
    % \makebox[\textwidth][c]{\includegraphics[width=0.85\textwidth]{figs/iprompt.pdf}}%    
    \vspace{-12pt}
    \caption{Overview of \method. \methods first proposes candidate prompts, then ranks them based on their performance as a prompt, then truncates and regenerates them. This entire process is repeated until performance stops improving.
    }
    \label{fig:method}
\end{figure}

\methods (\cref{fig:method}) is an iterative local search algorithm that alternates between three steps: (i) proposing candidate prompts, (ii) reranking candidate prompts, (iii) exploration.

\noindent
\text{(i) Proposal}:
Candidate prompts are generated by extending the zero-shot LLM generation. Given a data instance as a prefix, we sample a number of candidate prompts.
% \footnote{One could use either average suffix decoding or suffix decoding with a single sample. For computational efficiency, we use suffix decoding with only a single sample. We also add randomly decode the output rather than using beam-search, as our iterative procedure can recover from initially finding inaccurate candidates.}
The maximum length of each candidate is pre-specified and fixed. For example, in the add-two-numbers task (\cref{fig:method}), we may generate four candidates: \emph{\{Combine the numbers,  Return the output, Sum in order, Compute the output\}}.

\text{(ii) Reranking}:
Given candidates, the objective \cref{eq:objective} is evaluated for each candidate prompt $s$. The top few candidates which maximize the objective are kept, e.g. narrowing down the candidates to \emph{\{Combine the numbers, Sum in order\}}.

\text{(iii) Iterate with exploration}:
Each of the top candidates from reranking is truncated at a random position. These truncated candidates are used as a prefix when generating new candidate prompts via suffix decoding.
For example, we may randomly select the start of the previous candidates and fill in the endings: \emph{\{Combine the \blank, Sum \blank\}} $\to$ \emph{\{Combine the numbers, Combine both arguments, Sum the numbers, Sum all inputs\}}.

The algorithm is repeated until identifying a suitably strong $\hat s$, e.g. \emph{Sum the numbers}. Steps (i) and (iii) ensure that prompts remain fluent, while step (ii) improves the score of the prompts on the objective. Computationally, \methods only requires running inference on the pre-trained LLM, yielding a significantly lower memory requirement than methods such as AutoPrompt which require access to the LLM's gradients.

% \section{Results}
% \label{sec:results}
% In what follows, \cref{subsec:exp_details} gives experimental details.  In \cref{subsec:expl_recovery}, we explore \method's ability to rediscover a correct and fluent prompt on the variety of datasets laid out in \cref{tab:dataset_ovw}. In \cref{subsec:expl_recovery}, we explore whether \method's prompts remain effective when transferred to different LLMs. Finally, in \cref{subsec:sentiment_classification} we test the ability of \method to generate a useful prompt on a larger scale, by testing prompt generalization on real-world sentiment classification datasets.

\section{Experimental Setup}
\label{subsec:exp_details}

We consider two sets of experiments. First in \cref{sec:results}, we explore \method's ability to rediscover a correct and fluent prompt on the variety of simple instruction datasets (\cref{tab:dataset_ovw}, top)  with known answers. Experiments test the ability of the model to recover a known prompt while also remaining fluent in a way that generalize to human readers and to other language models. In \cref{sec:science} we apply \methods to scientific datasets (\cref{tab:dataset_ovw}, bottom).

\paragraph{Language Models} For the main set of experiments, we always generate prompts using GPT-J, a 6 billion parameter model~\citep{gpt_j}.  We restrict prompts to $\{$6,12$\}$ tokens for sentiment classification and 6 tokens for the remaining data collections in \cref{tab:dataset_ovw}.
For generalization experiments, alternative models are tested with the generated prompts including OPT and GPT-3~\cite{zhang2022opt,brown2020language}. See \cref{sec:exp_details_supp} for a full discussion of experimental details and \cref{subsec:rank_exps_extended} for experiments on more models (e.g. Galactica~\citep{taylor2022galactica})
% , Flan-T5~\citep{flan})
and more datasets.

\paragraph{Evaluation metrics}

We consider two types of evaluation: closeness to ground-truth and accuracy as a prompt. 
To measure closeness we use three metrics:
(1) Correct -- whether the generated explanation contains one of a set of problem-specific keywords.
% \footnote{The examples in each task do not directly contain any of the keywords; for example, when inferring the \textit{Add two numbers} task, the examples do not contain a plus sign or any synonyms of the word \textit{add} such as \textit{combine}.}
(2) MRR -- Mean reciprocal rank measuring the rank of the first task-correct prompt. Given a set of datasets $\mathcal{D} = \{\mathcal{D}_1, ..., \mathcal{D}_N\}$, we compute: $ \text{MRR} = \frac{1}{|\mathcal{D}|} \sum_{i=1}^{|\mathcal{D}|} \frac{1} {\text{rank}_i}$, where $\text{rank}_i$ is the one-indexed rank of the first correct explanation.
(3) Human -- The human evaluation scores between the top-generated explanation and a pre-specified groundtruth explanation, when instructed ``You are given a groundtruth description along with a generated one. On a scale of 1 (worst) to 5 (best), how interpretable and accurate is the generated description?''\footnote{Human evaluation scores are averaged over 4 PhD students in machine learning not affiliated with the study. }. The mean human evaluation score (ranging from 1 to 5) is normalized.

To measure generalization ability, we evaluate explanations based on accuracy as a prompt for other models.
% To evaluate similarity to the ground truth, 
% we score a ranked list of prompts based on mean reciprocal rank (MRR).
% Correctness could be determined in a problem-specific way, for example, by setting a threshold on minimum Levenshtein distance or BLEU score \citep{papineni2002bleu} with respect to the true explanation. 
% To measure generalization, we use the top-ranked string as a zero-shot prompt for a different language model, and evaluate whether that model is able to solve the task. 
Accuracy is computed following \cite{brown2020language,raffel2020exploring}: using exact matching with beam search, a beam width of 4, and a length penalty of $\alpha = 0.6$. 

For sentiment evaluation, we learn a prompt within the template \textit{Input: ``\$\{input\}''\{prompt\}}.\footnote{In initial experiments, we find that performance drops significantly when learning a prompt that comes \textit{before} the input.} We use \textit{positive} and \textit{negative} as positive and negative labels and require the LLM to rank the two options. Human-written prompts are adapted to this template from open-source prompts available through PromptSource \cite{bach2022promptsource}. % All experiments are run across 3 random seeds with the exception of human prompt accuracies which are averaged across 8 prompts.

% The aim is to find a dataset-specific prompt that can describe a particular sentiment classification setting. As there is no correct prompt for these datasets, evaluation is done on task accuracy.

\section{Results and Analysis}
\label{sec:results}

\subsection{Dataset explanation recovery}
\label{sec:expl_recovery}
% \paragraph{Recovering accurate dataset explanations} 
\cref{tab:prompt_accuracy} compares prompting methods across three diverse data collections.
The \textit{Human} evaluation scores are much higher for \methods than the baselines, suggesting that it finds prompts which are both accurate and human-interpretable.
% (omitting datasets for which no methods successfully retrieve an accurate prompt; see \cref{subsec:data_datails_supp}).
Similarly, the \textit{MRR} and \textit{Correct} scores show that \methods
% considerably increases the mean reciprocal rank %(\cref{sec:task_datasets})
considerably improves in finding accurate explanations.
See all generated explanations in \cref{tab:prompt_examples_full}.
% uggesting that \methods more effectively generates descriptions that accurately reflect the underlying data pattern.
% \footnote{We restrict generated prompts to 6 tokens (see a full discussion of experimental details in \cref{sec:exp_details_supp}).}

\begin{table}[t]
    \small
    \centering
    \caption{Performance for dataset explanation. Dataset from \cref{tab:dataset_ovw}~(1-3). Accuracy measured via (1) Human-evaluation (H, normalized \%), (2) Mean Reciprocal Rank across the collection (M) and (3) 1-best correctness (C, \%).
    % All experiments are on GPT-J 6B. 
    For all metrics, higher is better.
    }
    \hspace*{-0.3cm}\begin{tabular}{lccccc}
\toprule
  & \multicolumn{1}{c}{\textbf{iPrompt}} & \multicolumn{1}{c}{AutoPrompt} & \multicolumn{1}{c}{Suffix} \\
  & H  / M / C    & H  / M / C  & H  / M / C  \\
 \midrule
    Math & \textbf{60} / \textbf{0.69} / \textbf{60}& 25 / 0.14 / 13 & 20 / 0.08 / 03\\
  ANLI & \textbf{56} / \textbf{0.41} / \textbf{37} & 21 / 0.07 / 07& 25 / 0.06 / 01  \\
  Induction & \textbf{42} / \textbf{0.35} / \textbf{28} & 21 / 0.09 / 08 & 23 / 0.04 / 01\\
\bottomrule
\end{tabular}

    \label{tab:prompt_accuracy}
\end{table}
\begin{figure}[t]
    \centering
    % \vspace{-25pt}
    \includegraphics[width=0.9\columnwidth]{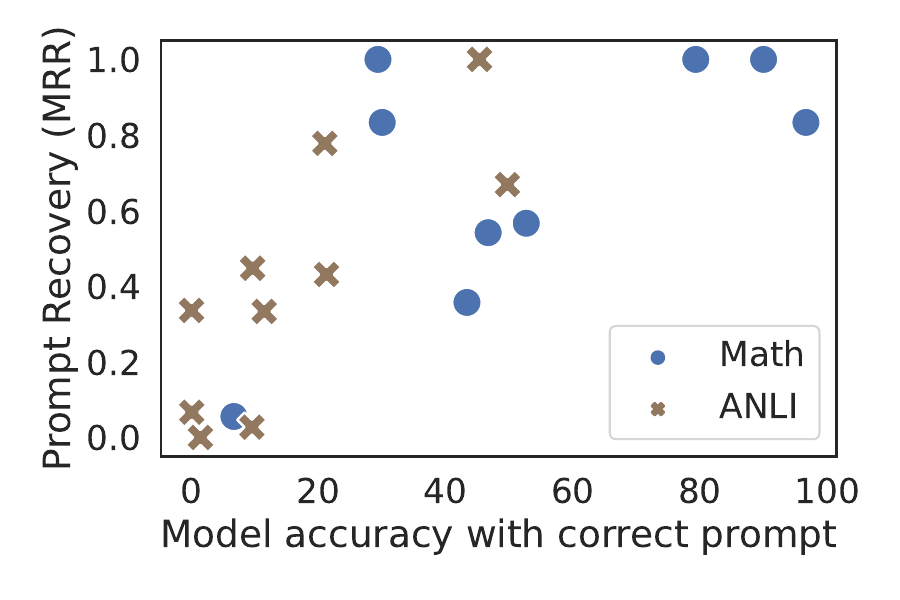}
    \vspace{-15pt}
    \caption{
    Comparison of model accuracy with correct prompt and iPrompt ability to find the correct prompt across each individual task (single-task MRR). Prompt recovery ability is dependent on the model's ability to perform the task.
    % Across a variety of datasets, iPrompt is able to recovery the correct prompt as long as the model is able to accurately complete the task.
    % across tasks for Math and ANLI datasets., plotted vs. model accuracy when prompt is provided as a prefix.
    }
    \label{fig:ablation_acc_vs_mrr}
\end{figure}
\begin{table}[t]
    \centering
    % \small
    \caption{Generalization accuracy (zero-shot) with the prompts generated with GPT-J as the LLM across different models.
    % \methods yields strong performance, usually improving over AutoPrompt despite maintaining interpretability, and sometimes performing close to the human-written prompt.
    % Numbers within 2\% of the top accuracy (excluding human-written prompts) for each model are shown in bold.
    }
    % \makebox[\textwidth][c]{
    \footnotesize
    \begin{tabular}{clc|ccc}
% {p{0.25\textwidth}p{0.1\textwidth} p{0.1\linewidth} p{0.1\linewidth}p{0.1\linewidth}p{0.1\linewidth}}
\toprule
{} & &  \makecell{Correct\\Prompt} & \textbf{iPrompt} &  AutoPrompt\hspace{-5pt}  &  \makecell{No\\prompt} \\
\midrule
& GPT-J 6.7B*           &          54.0 &    \textbf{51.5} &        41.6 &                 16.3 \\
%  \\
% & GPT 2.7B &           22.3 &     \textbf{23.5} &        \textbf{21.5} &             13.3 &       18.3 \\
% & OPT 2.7B &           17.7 &     \textbf{14.0} &         6.8 &              7.6 &       14.8 \\
\parbox[c]{0.3mm}{\multirow{2}{*}{\rotatebox[origin=c]{90} {Math}}} & OPT 6.7B &           12.7 &     \textbf{19.3} &        \textbf{18.9} &            8.4 \\
& GPT 20B  &           76.1 &     \textbf{54.4} &        23.2 &           8.5 \\
& GPT-3 175B &           76.0 &     \textbf{62.1} &        40.8 &               28.4 \\
\midrule
& GPT-J 6.7B*           &            9.0 &      \textbf{4.7} &         1.9 &               2.0 \\
% & GPT 2.7B &           13.1 &     14.1 &         8.5 &             13.4 &        5.2 \\
% & OPT 2.7B &            9.7 &      4.4 &         6.3 &              8.2 &        6.1 \\
\parbox[c]{0.3mm}{\multirow{2}{*}{\rotatebox[origin=c]{90} {ANLI}}}  & OPT 6.7B &           10.7 &      \textbf{6.7} &         4.7 &          \textbf{7.9} \\
& GPT 20B  &           31.0 &     \textbf{14.2} &         5.6 &           4.0 \\
& GPT-3 175B &           37.6 &     \textbf{11.7} &         2.7 &         7.7 \\
\bottomrule
\end{tabular}

    % }
    \label{tab:generalization}
\end{table}

To assess the best-case absolute accuracy of the approach, we note it is impossible for the approach to recover the prompt if the underlying LLM cannot solve the task.  \cref{fig:ablation_acc_vs_mrr} plots the prompt recovery performance (MRR) against the underlying LLM's accuracy (when using the groundtruth prompt) for each dataset. When the model can solve the task, \methods does well on recovery. However for many tasks 
the model has low accuracy even with the correct prompt, putting a ceiling on the performance of \method.
% at task with prompt as prefix across tasks from Math and ANLI.
% As the model's task accuracy increases, iPrompt is more likely to recover the correct prompt.
% For some ANLI datasets, both the MRR and task accuracy are very low.

\subsection{Generalization accuracy of prompts}
% \paragraph{Generalization accuracy induced by generated prompts.} 
Do prompts generated for a specific LLM still work when applied to a different model?
\cref{tab:generalization} shows the generalization accuracy when testing the prompts generated using GPT-J (\cref{tab:prompt_examples}) on different LLMs.
The prompts maintain effectiveness across most models.
For the Math datasets, the \methods prompts elicit improvement over the baselines and approach the accuracy of the correct prompt. For the ANLI datasets, all prompts induce poor performance.
Notably, the gap between \methods and AutoPrompt is larger for larger models (i.e. GPT 20B and GPT-3); this suggests that, by generating fluent prompts, \methods generates more generalizable descriptions.
% Human-written prompts still outperform the autoprompting methods on this task.
% In addition, when the prompt is evaluated using other model sizes (e.g. GPT-Neo 2.7 billion parameter model~\citep{gpt_neo}) or model architectures (e.g. OPT~\citep{zhang2022opt}), 
% Occasionally, \methods is even able to outperform the performance of human-written prompts.

\begin{table}[tt]
    \centering
    \small
    \caption{Zero-shot accuracy on sentiment classification datasets: SST-2, Rotten Tomatoes, IMDB, and the Financial Phrasebank ~\citep{socher2013recursive,Malo2014GoodDO,PangLee2005}.
    Generation with GPT-J 6B and evaluation on both on the original GPT-J model
     and 
     % testing generalization to
    GPT-3 (\texttt{text-davinci-002}).  
    % Values are averaged over three random seeds for prompt-generation; 
    Errors are standard errors of the mean.
    }
    \label{tab:sentiment}
    % \makebox[\textwidth][c]{
    \vspace{1 mm}
\begin{tabular}{llcccc}
% {p{0.25\textwidth}p{0.16\textwidth} p{0.16\linewidth} p{0.16\linewidth}p{0.16\linewidth}p{0.16\linewidth}}
% \begin{tabular}{llll}

\toprule {} & & \makecell{Human-\\written} & \textbf{iPrompt} & AutoPrompt & \makecell{No\\prompt} \\
\midrule
\parbox[c]{1mm}{\multirow{4}{*}{\rotatebox[origin=c]{90} {GPT-J}}}
& FFB & 27.0 $\pm$ 1.9 &\textbf{ 79.3 $\pm$ 2.1} & 74.0 $\pm$ 9.1 & 47.5 \\
& RT & 58.9 $\pm$ 3.1 & \textbf{84.8 $\pm$ 0.9} & 73.0 $\pm$ 4.8 & 59.2 \\
& SST-2 & 58.4 $\pm$ 2.8 & \textbf{86.7 $\pm$ 1.0} & 76.7 $\pm$ 3.9 & 60.9 \\
& IMDB & 66.0 $\pm$ 3.2 & \textbf{87.9 $\pm$ 1.4} & 86.7 $\pm$ 1.2 & 58.6 \\
\midrule
\parbox[c]{1mm}{\multirow{4}{*}{\rotatebox[origin=c]{90} {GPT-3}}}
& FFB & 39.6 $\pm$ 1.6 & \textbf{57.2 $\pm$ 6.9} & 28.2 $\pm$ 3.1 & 39.1  \\
& RT & \textbf{82.7} $\pm$ 3.3 & 77.4 $\pm$ 2.8 & 57.8 $\pm$ 3.5 & 54.8\\
& SST-2 & \textbf{90.5} $\pm$ 3.9 & 82.4 $\pm$ 2.3 & 61.8 $\pm$ 7.0 & 58.4 \\
& IMDB & 75.6 $\pm$ 3.3 & \textbf{86.6 $\pm$ 1.1} & 70.0 $\pm$ 6.5 & 66.2  
\\
\bottomrule
\end{tabular}

% \begin{tabular}{llcccc}
% % {p{0.25\textwidth}p{0.16\textwidth} p{0.16\linewidth} p{0.16\linewidth}p{0.16\linewidth}p{0.16\linewidth}}
% % \begin{tabular}{llll}
% \toprule {} & & \makecell{Human-\\written} & \textbf{iPrompt} & AutoPrompt & \makecell{No\\prompt} \\
% \midrule
% \parbox[c]{1mm}{\multirow{4}{*}{\rotatebox[origin=c]{90} {GPT-J}}} &
% FFB & 24.3 & \textbf{62.4 $\pm$ 0.1} & 6.8 $\pm$ 2.9 & 0.0 \\
% & RT & 44.4 & \textbf{70.5 $\pm$ 1.4} & 57.1 $\pm$ 3.4 & 0.0 \\
% & SST-2 & 53.6 & \textbf{82.8 $\pm$ 1.9} & 40.0 $\pm$ 7.9 & 0.0 \\
% & IMDB & \textbf{32.5} & 21.3 $\pm$ 9.3 & 12.1 $\pm$ 0.9 & 3.5 \\
% \midrule
% % Tweet (Hate) & 16.1 & 93.9 & 78.3 & 0.0 \\
% \parbox[c]{1mm}{\multirow{4}{*}{\rotatebox[origin=c]{90} {GPT-3}}} & FFB & 54.1 & \textbf{65.0} & 2.7 & 0.4 \\
% & RT & \textbf{58.6} & 52.5 & 37.5 & 0.9 \\
% & SST-2 & 60.4 & \textbf{83.6} & 5.2 & 0.6 \\
% & IMDB & \textbf{79.0} & 1.3 & 0.9 & 1.1 \\
% \bottomrule
% \end{tabular}
    % }
\end{table}

\begin{table*}[t]
    \centering
    \scriptsize
    % \footnotesize
    \caption{
    Examples of generated explanations by \methods and AutoPrompt.
    See all prompts in \cref{tab:prompt_examples_full}.
    % In contrast, AutoPrompt (and other discrete prompting methods) generate seemingly nonsensical prompts (see \cref{tab:prompt_examples_full}).
    }
    \renewcommand{\arraystretch}{1.25} % stretch the spacing in this table a tiny bit
\begin{tabular}{r>{}p{0.31\textwidth} p{0.27\textwidth} p{0.35\textwidth} }
\toprule
 & Human-written prompt & \textbf{iPrompt} & AutoPrompt\\
\midrule
 % \parbox[c]{3mm}{\multirow{5}{*}{\rotatebox[origin=c]{90} {\makecell{\cblue{(i)} Describe\\unknown function}}}} 
  \parbox[c]{0.5mm}{\multirow{3}{*}{\rotatebox[origin=c]{90}{Math}}} 
 & Return the sum of the inputs & Create a function named `sum & >:Returns Adding togetherFont accomplish\\
 & Return the square of the input & Input number and return its square	& Cal impl qApplySquare fiat\\
  & Differentiate between prime/non-prime integers & Are these pairs of integers prime & ropheospels\&\& Norestricted 	\\
 \midrule
 \parbox[c]{0.5mm}{\multirow{6}{*}{\rotatebox[origin=c]{90} {ANLI}}}
 & Differentiate vegetarian/non-vegetarian foods & Are you a vegetarian?	& compliedthe whether methamphetamine provided comp	\\
 & Differentiate the subject in a sentence based on gender & Predict the gender (F =	& < endoftext > -> M Fundamental FG Fav	\\
 & Return a synonym & what is a synonym for	& Word termOn English meanings		\\
 & Translate english to spanish & please write English meaning in Spanish & the ththebb volunt\\
 & Return a country's capital city &  Which city is the capital and	& Ang Suppose AUTHthe beh Assassins \\
 % \vspace{1mm} % spacing  at bottom of box 
\midrule
\parbox[c]{0.5mm}{\multirow{3}{*}{\rotatebox[origin=c]{90} {Sentiment}}} & What is the sentiment expressed by the reviewer for the movie? & Describe what it is about this film has caused it & Pap Azerb Saiyan Forean Talatar Yemeni IndBloomberg receiveda \\
 &  How does the author of the news headline feel? & $<$input$>$ neutral$>$ The result was due to: "  & Fur resultolandgroundur augmented= \\
 % \vspace{4mm}
\bottomrule
\end{tabular}
    % }
    \label{tab:prompt_examples}
\end{table*}

\cref{tab:sentiment} shows results on the sentiment analysis datasets. 
As prompts for GPT-J, \methods outperforms not only AutoPrompt, but also the manually-written prompt on all four datasets. Interestingly, the average performance of human-written prompts on GPT-J is very low, unlike the prompts generated by \method. This indicates that models at 6B parameter scale may be brittle to the choice of prompt, even among a set of reasonable options, and \methods (and to an extent, AutoPrompt) is able to discover how to phrase prompts so that models of this scale can complete the task. % chandan let me know if this part makes sense

When sentiment prompt generalization is tested on GPT-3, we find that \methods prompts outperform human-written prompts on two of the four datasets. When tested on GPT-3, \methods prompt \textit{To summarize this review! :} outperforms all PromptSource IMDB prompts that use the same verbalizer (\textit{positive}/\textit{negative}).
When its prompts are tested on GPT-3, baseline AutoPrompt only slightly outperforms testing with no prompt at all.

% \subsection {Sentiment classification}
% % Investigating \methods in sentiment classification
% \label{subsec:sentiment_classification}

 \cref{tab:prompt_examples} shows the top-ranked explanation generated by each method for selected datasets.
\methods often finds an explanation that is indicative of the underlying relationship, even if the phrasing is not perfect.
For example, for the \textit{add two numbers} dataset, it finds \textit{Create a function named `sum}.
% For difficult datasets, the \methods string sometimes simply returns the classes of the output (e.g. \textit{yes or no?}) rather than capturing the underlying relationship.
The prompts found by \methods also read as fairly fluent strings compared to AutoPrompt, which produces an incoherent set of tokens.
% For example, for the \textit{add two numbers} dataset, AutoPrompt finds the prompt \textit{>:Returns Adding togetherFont accomplish} and for the \textit{vegetarian} dataset, it finds the prompt \textit{compliedthe whether methamphetamine provided comp}.

% This suggests that \method's ability to recover prompts will improve as underlying LLMs improve.
% This explains the poor performance of GPT-J on some tasks,
% and also suggests that \methods's ability to recover prompts will improve as underlying LLMs improve.
% including the 'exp' task and several from ANLI: because the tasks were too hard for the model, iPrompt was unable to discriminate between good and bad prompts.

\subsection{Model ablations}
\label{subsec:ablations}
We run ablation experiments to analyze the three steps of iPrompt: (1) Proposal, (2) Reranking, and (3) Iteration.
% . Unless stated otherwise, all experiments use 
% We use the GPT-J model
% and are averaged over 3 random seeds
% and the Math/ANLI datasets to
% learn a 6-token prompt and
we use the Math and ANLI datasets and run on a maximum of 5000 data points using 5 shots in context for prompt generation. % All experiments are averaged over 3 random seeds.

\begin{table}[t]
    \centering
    \small
    \caption{Algorithmic ablations for each stage of \method. Gives prompt recovery (MRR) achieved by ablating each stage. Averaged over 3 random seeds.}
    \begin{tabular}{llll}
    \toprule
        & & \multicolumn{2}{c}{MRR}\\
         & & Math & ANLI \\
    \midrule
     & \textbf{iPrompt} & \textbf{0.557} & \textbf{0.278} \\
    \midrule
    (1) Proposal &  w/o inputs+outputs & 0.400 & 0.015 \\
       &  w/o inputs & 0.463  & 0.244 \\
    &    w/o outputs & 0.539  & 0.255 \\
    \midrule
    (2) Reranking &  w/ in-context examples & 0.071 & 0.152 \\
    \midrule
    (3) Iteration & No iteration & 0.075 & 0.050 \\
    \bottomrule
\end{tabular}
    \label{tab:ablation}
\end{table}

% \textbf{(1) Proposals help model search.}
\textbf{(1) \textit{Proposals}  are partially guided by examples.}
During the proposal stage, \methods prefixes potential prompts with dataset examples. \cref{tab:ablation} considers variants of this stage that remove input and output examples during the proposal stage. Note the system still has access to the full examples during the reranking stage. We find the system can achieve decent performance on Math simply by iterating. However for ANLI, the model needs to at least see the inputs/outputs during the proposal in order to find accurate prompts. 

% These ablations indicate that the reranking step can succeed even when proposals are noisy. 
% in fact iPrompt achieves an MRR of $0.40$ on the Math datasets by simply generating unconditional prompts, and iteratively reranking them.

\textbf{(2) \textit{Reranking} zero-shot recovers better prompts.} 
\methods uses zero-shot accuracy to rank prompts. As we have examples of the task, we could instead use in-context few-shot prompting for ranking. Prior work suggests that prompt wording is less influential as the number of in-context examples increases \cite{webson2021prompt}. \cref{tab:ablation} shows that using these examples in-context for reranking does, in fact, considerably hamper prompt recovery.
% On both Math and ANLI, when prompts are re-ranked based on few-shot instead of zero-shot performance, MRR drops significantly.
% – from $0.557$ to $0.071$ for Math and from $0.278$ to $0.152$ for ANLI.
% We conclude that it is important to discriminate prompts in the single-shot rather than few-shot setting.
We further find that the LLM used for reranking is more important than the LLM used for proposals (see Appendix \cref{fig:ablation_disc_heatmap}).

\textbf{(3) \textit{Iteration} improves performance}
Finally, \cref{tab:ablation} shows that without multiple iterations, performance drops nearly to zero (\cref{fig:ablation_loss_convergence} shows more details on loss as a function of iterations).

\section{Scientific investigations with \method}
\label{sec:science}

We now investigate whether \methods can explain patterns in scientific datasets.
Specifically, we analyze the Galactica model~\cite{taylor2022galactica} with 6.7 billion parameters. We query whether it can describe differences in datasets of chemical compounds and protein sequences before investigating a neuroscience problem.

\paragraph{Toxic chemical compounds} We first ask whether \methods can explain the difference between two groups of chemical compounds with a known difference.
We use the Tox21 dataset~\cite{richard2020tox21} which contains toxicity measurements on 12 biological targets.
For each of the 12 biological targets, we search for a prompt that differentiates compounds that are toxic to the target (positive) from those which are not toxic to any of the targets (negative).
We use 100 positive/negative examples for each biological target and format each input with the text \textit{Here is a compound:\textbackslash n [Compound Name]\textbackslash n Answer:} followed by \textit{Yes} for a positive compound and \textit{No} for a negative one.
iPrompt is run for a single epoch with 5 shots in each example.

% Prefix: ``Answer Yes if the compound is``
Ideally, the elicited prompt would mention toxicity.
\cref{tab:tox21} shows results for whether the elicited prompts contain the substring \textit{tox}, both in terms of MRR and top-prompt correctness.
iPrompt often finds an accurate prompt: one representative example is: \textit{Answer yes if the compound is toxic, and Otherwise answer NO}.
To ensure that this substring is not simply a popular completion for the language model, we compare against a baseline which runs iPrompt using Galactica proposals from empty inputs/outputs and reranking with Galactica; over 36 random seeds, \textit{tox} appears in any generated prompt.

\begin{table}[t]
    \centering
    \small
    \caption{
    iPrompt performance at recovering prompts for toxic chemical compounds.
    Tox21 results are averaged over 12 datasets with 3 random seeds each.
    Null data is averaged over 36 random seeds.
    Error bars are standard error of the mean.
    }
    \begin{tabular}{llr}
\toprule
{} &          iPrompt &  Baseline \\
\midrule
MRR                    &  0.83 $\pm$ 0.04 &        0.0 \\
Top-prompt correctness &  0.67 $\pm$ 0.08 &        0.0 \\
\bottomrule
\end{tabular}
    \label{tab:tox21}
\end{table}

\paragraph{Differentiating protein sequences}
We turn to whether \methods can explain the differences between two groups of proteins.
We use protein sequences and keywords from
Swiss-Prot~\citep{bairoch1991swiss} (a high-quality subset of Uniprot~\citep{uniprot2015uniprot}) to construct two datasets:
each dataset contains two groups of proteins, which are differentiated based on their keywords.\footnote{We search for reasonably popular but non-cooccuring keywords in the proteins; see details in \cref{fig:uniprot_keyword_coocurences}}
The first dataset, which we call \textit{Cyto}, has proteins with either the keyword \textit{Cytoplasm} or \textit{Membrane}.
The second dataset, which we call \textit{Binding}, has proteins with either the keyword \textit{RNA-binding} or \textit{ATP-binding}.
Each group is randomly downsampled to 100 proteins and \methods is run with the same hyperparameters as when finding chemical compounds.

We make this problem more challenging by feeding the model the raw protein sequence (not the protein name) which ranges from hundreds to thousands of amino acids.
Each input is presented with the following text:
\textit{Here is a protein sequence:\textbackslash n [Protein Sequence]\textbackslash n Answer:} followed by \textit{Yes} for a one group and \textit{No} for the other.
\cref{tab:uniprot} shows results for identifying whether the elicited prompt contains one of the relevant keywords for each dataset (e.g. \textit{Cytoplasm}).
Despite the difficult input format, the correct keywords are successfully identified for both the \textit{Cyto} and \textit{Binding} datasets better than for the \textit{Baseline} (which again contains empty inputs).

\begin{table}[t]
    \centering
    \small
    \caption{
    iPrompt performance at differentiating protein sequences.
    For both the \textit{Cyto} and \textit{Binding} datasets, the correct keywords are succesfully identified better than for the Baseline.
    Results are averaged over 12 random seeds; error bars are standard error of the mean.
    }
    \begin{tabular}{llll}
\toprule
{} &             iPrompt (Cyto) &     iPrompt (Binding) &        Baseline \\
\midrule
MRR                &   0.2 $\pm$ 0.08 &  0.08 $\pm$ 0.04 &  0.03 $\pm$ 0.01 \\
Recall @ 5  &  0.25 $\pm$ 0.13 &  0.17 $\pm$ 0.11 &  0.05 $\pm$ 0.05 \\
Recall @ 20 &  0.83 $\pm$ 0.11 &  0.33 $\pm$ 0.14 &  0.23 $\pm$ 0.09 \\
\bottomrule
\end{tabular}

    \label{tab:uniprot}
\end{table}

\paragraph{Scientific investigation into an fMRI natural language dataset}
% \label{subsec:fmri}
We now explore using \methods in a simple neuroscience experiment.
A central challenge in neuroscience is understanding how and where semantic concepts are represented in the brain.
A recent seminal study~\citep{huth2016natural} explores this question by investigating where different natural language categories are represented in the human neocortex.
Specifically, the authors collect functional MRI (fMRI) responses as human subjects listen to hours of narrative stories.
They then build a predictive model of these responses for each voxel (i.e. a small region in space) in the brain, which takes as input the words contained in the stories (and other features).
To interpret these individual voxel models,
they cluster the words in the narrative stories into 12 groups and manually annotate them, resulting in 12 categories, such as \textit{tactile}, \textit{visual}, and \textit{professional}.
Finally, they view the spatial mapping of these 12 concepts (projected onto low dimensions) across the brain using their individual voxel models.

We revisit a small piece of this study's analysis through the lens of \method.
Specifically, we ask whether \methods could generate plausible categories that are well-represented across the brain but differ from the manually identified 12.
We fit a predictive model for each voxel, following the pipeline of the original study
% which involves many steps such as aligning the input stories with the recorded responses and accounting for delayed responses in fMRI data
(details in \cref{sec:fmri_supp}).
We then use the resulting models to identify a list of the top-15 words which most excite each voxel.
For example, the top-15 words that excite the best-predicted voxel are:
\textit{sheet, edges, diameter, strips, cardboard, copper, steel, colored, coloured, leaf, wire, cap, paper, shaped, tin}.
To identify a plausible semantic category, we construct a template string as follows: 
% \begin{addmargin}[2em]{2em}% 1em left, 2em right
% \begin{center}
% \begin{split}
% \label{words_list}
\textit{The following list of words all belong to the same semantic category: \blank\textbackslash n\textbackslash
n sheet, edges, ..., shaped, tin}.
% \end{center}
% &\text{The semantic category they all belong to, in one word, is \blank''}
% \end{addmargin}
% \end{split}
We then use \methods (again with a GPT-6B parameter model) to generate a category by filling in the blank (restricted to a single token).
To make \methods more effective, for each voxel we use \methods on a set of examples consisting of 15 permutations of the top-15 words, allowing finding patterns that are not overly sensitive to the word-ordering.

\begin{figure}[t]
    \centering
    % \vspace{-25pt}
     % \makebox[\textwidth][c]{\includegraphics[width=1.0\textwidth]{figs/flatmap.pdf}}%
    \adjincludegraphics[width=0.9\columnwidth,trim={{.49\width} {.03\height} 0 {.03\height}},clip]{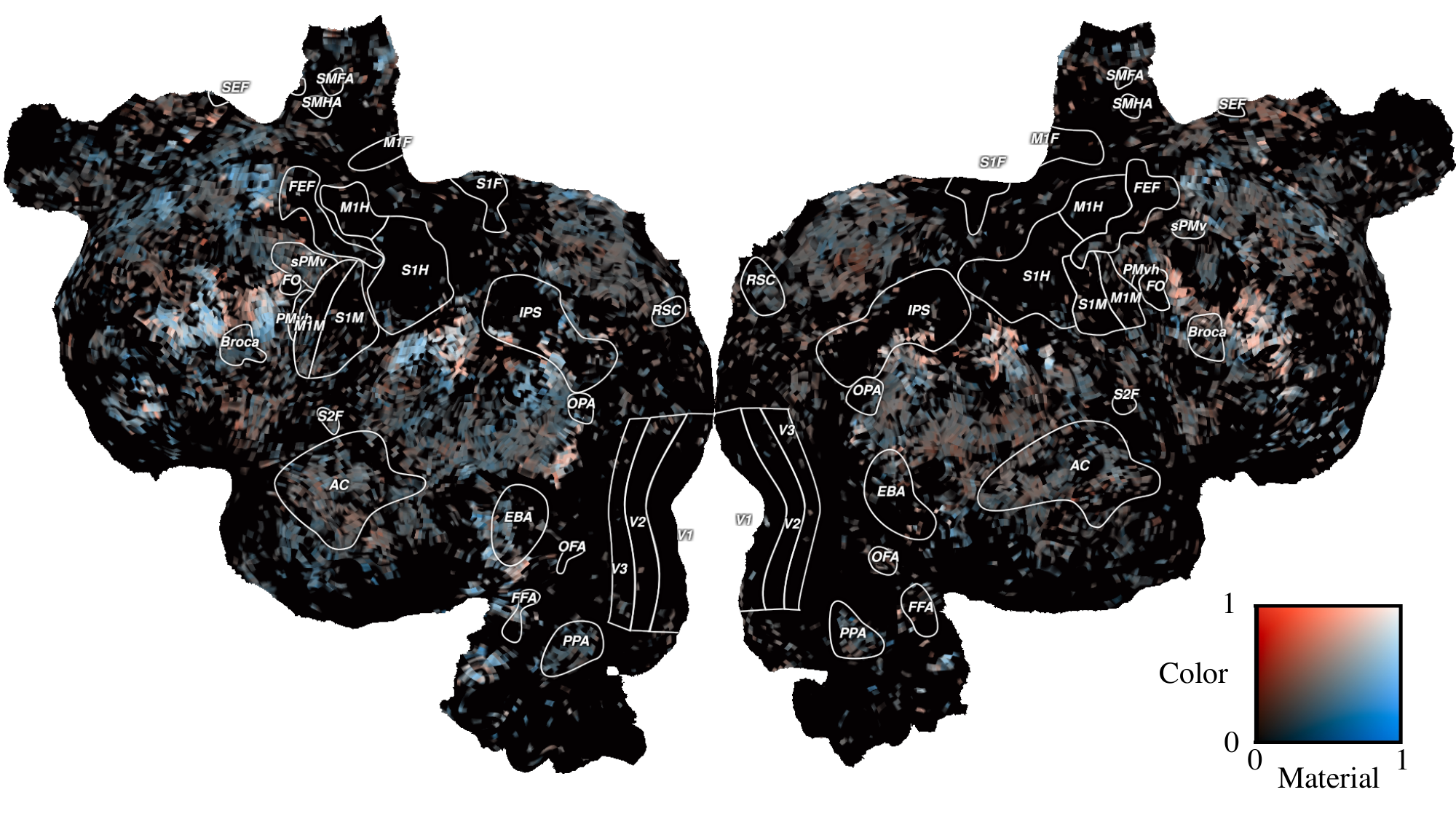}
    \vspace{-11pt}
    \caption{Representations of the \method-elicited concepts \textit{material} (blue) and \textit{color} (red) across the surface of the neocortex are spatially clustered and smooth.
    % Each point shows the inferred voxel selectivity of the categories \textit{material} and \textit{color}, measured by an LLM.
    Only the top 10,000 best-predicted voxels are shown, remaining voxels are shown in black.
    Only the right hemisphere is shown (see both hemispheres, which show consistent smoothness in \cref{fig:flatmap_supp}).
    % Plotted with pycortex~\citep{gao2015pycortex}.
    }
    \label{fig:flatmap}
\end{figure}

Given the top categories for each voxel, we analyze the mapping of recurring categories across the neocortex.
We aggregate the top-15 inferred categories\footnote{We apply stemming and remove stopwords before choosing the best categories.} over the top-15 best-predicted voxels and find that
the most frequently inferred categories are:
% \begin{addmargin}[2em]{2em}% 1em left, 2em right
% \begin{equation}
% \label{offset:top_categories}
\texttt{material, color, surface, text, \& fabric}.
% \end{equation}
% \end{addmargin}
Interestingly, these are sensible quantities that different voxels could reasonably be selective for.
% We now spatially map each of these identified categories (e.g. \textit{material}) across the 10,000 best-predicted voxels in the brain by querying the response of the constructed voxel-wise models as in \cite{huth2016natural}.
We spatially map each of these identified categories (e.g. \textit{material}) across the 10,000 best-predicted voxels by using the LLM in a second way.
For each voxel, we condition the LLM (again GPT-6B) on the top-15 words list, and evaluate the predicted probability for each category, i.e. \textit{The following list of words all belong to the same semantic category:
sheet, edges, ..., shaped, tin The semantic category they all belong to, in one word, is \blank}.
The higher this predicted probability, the more selective we infer that a voxel is for the category.
\cref{fig:flatmap} shows these predicted probabilities for the top-two inferred categories (\textit{material} and \textit{color}) across the cortex of a human subject.

While there is no groundtruth for this semantic map, one noteworthy feature of the resulting map is that it is spatially smooth (quantitatively, \cref{fig:concepts_clustered} shows that the variance of the map among neighboring pixels is significantly lower than we would expect by shuffling the map's values).
This is non-trivial, as nowhere in the modeling process was spatial information incorporated:
each voxel was modeled independently and the displayed prediction was queried independently.
We expect the underlying map to be smooth,
both due to local connectivity in brain regions and also because the BOLD signal measured by fMRI does not have perfect spatial resolution.
Thus, the fact that our inferred map is smooth suggests that
(i) something about these categories is genuinely captured by the representation in the human brain,
and (ii) that the \methods approach was able to reflect at least some of it.
Beyond the two categories shown, the five categories generated by \methods exhibit spatial smoothness across the neocortex (\cref{fig:concepts_clustered}).

% \subsection{Interpretable prompt-based clustering}
% \label{subsec:clustering}
% In interpretable prompt-based clustering, we seek to cluster individual data examples,
% again given a template and an LLM $f$.
% We do this via a simple two-step procedure:
% we first compute the probability vector $P(\text{next}; f, \textit{template})$ for each example in the dataset.
% We then cluster these vectors, simply using k-means after applying dimensionality reduction (principal component analysis).
% In the second stage, we generate a full description for each cluster by following the same procedure as before (using average-output decoding) using the examples within each cluster.
% For this two-stage approach to work effectively, we choose a template that yields an important distinction in the first token of $\hat s$,
% although for more complex cases one may need to instead cluster the assigned probability over descriptions of larger length.

% \begin{figure}[t]
%     \centering
%     \includegraphics[width=0.6\textwidth]{figs/clustering.png}
%     \caption{Pairwise clustering results. \cs{Need to clean this up}}
%     \label{fig:clustering}
% \end{figure}

\section{Conclusion and Discussion}
\label{sec:discussion}

\methods makes a meaningful step towards finding natural language prompts that are both accurate and  human-interpretable. We show this method can be used to recover dataset descriptions, produce transferable prompts, and provide explanations for experimental data. 
% Notably, \methods allows the flexible use of domain knowledge and improves as the underlying LLM used to generate prompts improves.
% Nevertheless, the search algorithms used in this work are computationally intensive;
% and fail to recover descriptions of complex datasets.
% Besides algorithmic improvements, 
% Future work could explore algorithmic variants that make \methodlongs more efficient and accurate, such as imbuing the search space with the ability so search over textual programs that can be used as input prompts~\cite{beurer2022prompting}.
One future direction could elicit \textit{targeted} information from data 
% incorporating human knowledge
via the use of a \textit{template}. 
% incorporate more information by using a \textit{template} to elicit more targeted information from data, using human prior knowledge.
For example, one may use \methods to extract feature importance
% rather than searching for a dataset description, we may search for a description of feature importance
by prepending the learned prompt with the string ``To get the answer from the inputs, the most important inputs are \blank''.
As another example, in a scientific study such as the fMRI study in \cref{sec:science}, a scientist interested in a particular topic (e.g. \textit{fear}) may investigate that particular topic by making a more specific template (e.g. \textit{How are these words related to the concept of ``fear''}?).
% With a sufficiently powerful LLM, this could potentially enable answering  targeted questions from opaque data.

While we focus on text, \methods could be applied generally settings where an LLM performs well.
For example, in computer vision, an interpretable autoprompt may look like a mask of an image, and in vision-language models, an interpretable prompt may be a description of a vision task, e.g. \textit{find the largest shape in this image}.
% This work is just the beginning of an interesting line of inquiry,
% and
% We hope \methods can help step towards the goal of yielding understanding from complex data with LLMs.

\section*{Acknowledgements}

AR is supported by NSF CAREER 2037519, NSF 1704834, and a Sloan Fellowship.
JM is supported by Weill Cornell Medicine.
Thanks to Wenting Zhao and Woojeong Kim for comments on drafts of this paper and to Jeevana Priya Inala, Xin Wang, Baolin Peng, Michel Galley, and Hao Cheng for interesting discussions related to the work.
We would also like to thank the authors of \cite{huth2016natural} for making their data publicly available.
\FloatBarrier

{
    \small
    % \bibliography{refs}
    \bibliographystyle{icml2023}

}

%%%%%%%%%%%%%%%%%%%%%%%%%%%%%%%%%%%%%%%%%%%%%%%%%%%%%%%%%%%%%%%%%%%%%%%%%%%%%%%
%%%%%%%%%%%%%%%%%%%%%%%%%%%%%%%%%%%%%%%%%%%%%%%%%%%%%%%%%%%%%%%%%%%%%%%%%%%%%%%
% APPENDIX
%%%%%%%%%%%%%%%%%%%%%%%%%%%%%%%%%%%%%%%%%%%%%%%%%%%%%%%%%%%%%%%%%%%%%%%%%%%%%%%
%%%%%%%%%%%%%%%%%%%%%%%%%%%%%%%%%%%%%%%%%%%%%%%%%%%%%%%%%%%%%%%%%%%%%%%%%%%%%%%
\newpage
\appendix
\onecolumn
\appendix
\setcounter{table}{0}
\setcounter{figure}{0}
\renewcommand{\thefigure}{A\arabic{figure}}
\renewcommand{\thetable}{A\arabic{table}}

\section{Appendix}
\label{sec:appendix}

\subsection{Sentiment classification results}
\label{app:sentiment-prompts}

\begin{figure}[H]
    \centering
    \includegraphics[width=0.7\textwidth]{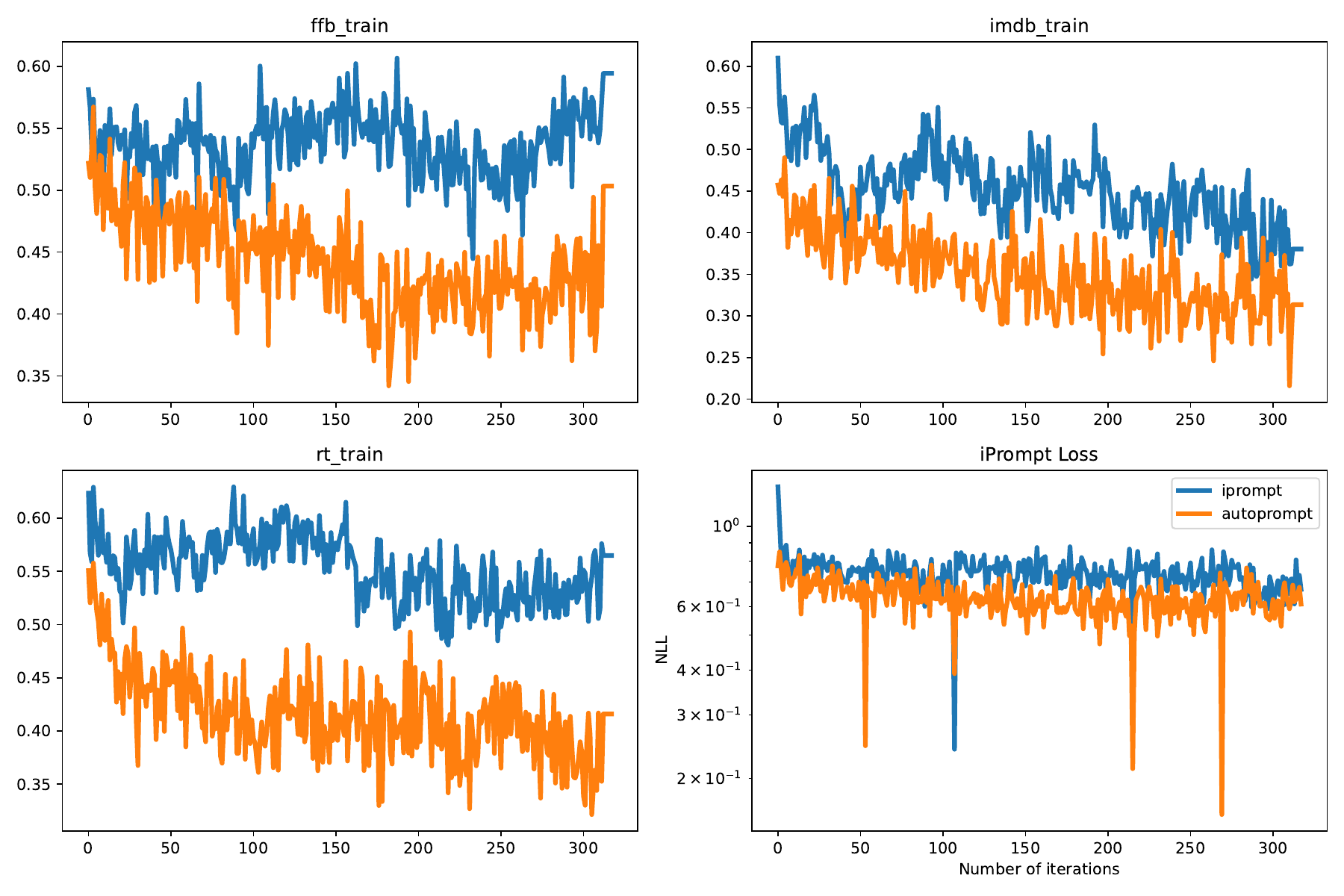}
    \caption{Loss plots for methods across sentiment analysis datasets, showing AutoPrompt and \methods across three random seeds.}
    \label{fig:sentiment_loss_plots}
\end{figure}

Table \ref{tab:sentiment-prompts} shows the best prompt produced by each method for each sentiment dataset. \methods often learns to recreate significant examples from the dataset, as a prompt. Figure \ref{fig:sentiment_loss_plots} shows loss across training step for each method and dataset, across three random seeds. We see that AutoPrompt often finds a prompt with slightly lower loss on the training data, although its prompts lead to worse generalization, as reported in \cref{tab:sentiment}. Each training step represents a single word swap (in the case of AutoPrompt) or the truncation and generation of a new prefix (in the case of \method).

Different from the other experiments in this paper, for sentiment classification, we initialize AutoPrompt with random tokens instead of all \textit{the}, as we find AutoPrompt fails to find an effective solution for longer prefix lengths when all tokens are initialized to \textit{the}. To accommodate for a complex input-output relationship, we test prompts of length 12 as well as length 6.

Accuracy is measured on the test set when available; otherwise, it is measured on a held-out 25\% of the train set.

\begin{table}[H]
    \centering
    \small
    \caption{Best-of-three prompts generated by each method on sentiment classification datasets. (Human-written prompts are best-of-eight and take from PromptSource \cite{bach2022promptsource}).}
    \label{tab:sentiment-prompts}
    \makebox[\textwidth][c]{
        \begin{tabular}{lll}
\toprule
 Task & Method & Prompt  \\
 \midrule
\multirow[c]{4}{*}{Financial phrasebank} & AutoPrompt &  Fur resultolandgroundur augmented \\
 & Human-written prompt & How does the author of the news headline feel?  \\
 & iPrompt &  $<$input$>$ neutral$>$ The result was due to: " \\
 \midrule
\multirow[c]{4}{*}{IMDB} & AutoPrompt &  uclear  cend Koretravel NAACP curses SicAstings production received \\
 & Human-written prompt & The movie review in negative/positive sentiment is:  \\
 & iPrompt &  This movie needs to be put up on my profile as my	
 \\
 \midrule
\multirow[c]{4}{*}{Rotten Tomatoes} & AutoPrompt &  Whether\{\{ anotherath<|endoftext|> how \\
 & Human-written prompt & What sentiment does the writer express for the movie?  \\
 & iPrompt & what words would you try to add to help you express that \\
 \midrule
\multirow[c]{4}{*}{SST-2} & AutoPrompt &  BryceSpecificallyWASHINGTONRatedam	 \\
 & Human-written prompt &  What is the sentiment expressed in this text? \\
 & iPrompt & It is clear from the sentence that all three actors have something \\
 \bottomrule
\end{tabular}

    }
\end{table}

% \begin{figure}[H]
%     \centering
%     \includegraphics[width=1.0\textwidth]{figs/sent_vary_test_dset.pdf}
%     \caption{Accuracy when testing sentiment-classification prompts on different test sets.
%     Testing on financial phrasebank (FFB) yields much lower performance, but testing on the other tree datasets, which are all movie reviews, yields reasonable performance.
%     iPrompt tends to find prompts that generalize better to new datasets.
%     All experients using GPT-J-6B.}
%     \label{fig:sent_vary_test_dset}
% \end{figure}

\subsection{Data/model details}
\label{subsec:data_datails_supp}

\begin{table}[H]
    \centering
    \scriptsize
    \caption{Details for each dataset. For details on \textit{Instruction induction}, see \cite{honovich2022instruction} and for details on \textit{Distribution differences}, see \cite{zhong2021adapting}.}
    \makebox[\textwidth][c]{
    \begin{tabular}{p{0.24\textwidth} p{0.07\linewidth} p{0.35\linewidth} p{0.35\linewidth}}

\toprule
                                                  Task name &  Samples &                                                                                                                                                                                                                                                                                                                                                                                                                                                                                                                                                                                                                                                                                                                                                                                                                                                                                                                                                                                                                              Description &                                                                                                                                                                                                                                     Example \\
\midrule
                                              fibonacci\_one &       10 &                                                                                                                                                                                                                                                                                                                                                                                                                                                                                                                                                                                                                                                                                                                                                                                                                                                                                                                                                                                       Given an input x, return the xth fibonacci number. &                                                                                                                                                                                          Given the input x is 8, the output f(x) is 21.\textbackslash n\textbackslash n \\
                                                 double\_one &       10 &                                                                                                                                                                                                                                                                                                                                                                                                                                                                                                                                                                                                                                                                                                                                                                                                                                                                                                                                                                                                            Given an input x, return 2*x. &                                                                                                                                                                                          Given the input x is 6, the output f(x) is 12.\textbackslash n\textbackslash n \\
                                                    exp\_one &       10 &                                                                                                                                                                                                                                                                                                                                                                                                                                                                                                                                                                                                                                                                                                                                                                                                                                                                                                                                                                                                Exponentiate the input to get the output. &                                                                                                                                                                                     Given the input x is 8, the output f(x) is 2980.96.\textbackslash n\textbackslash n \\
                                                 square\_one &       10 &                                                                                                                                                                                                                                                                                                                                                                                                                                                                                                                                                                                                                                                                                                                                                                                                                                                                                                                                                                                                      Square the input to get the output. &                                                                                                                                                                                           Given the input x is 2, the output f(x) is 4.\textbackslash n\textbackslash n \\
                                                  first\_two &      100 &                                                                                                                                                                                                                                                                                                                                                                                                                                                                                                                                                                                                                                                                                                                                                                                                                                                                                                                                                                                                          Return the first of the inputs. &                                                                                                                                                                                       Given the input numbers 7 and 8, the answer is 7.\textbackslash n\textbackslash n \\
                                                    add\_two &      100 &                                                                                                                                                                                                                                                                                                                                                                                                                                                                                                                                                                                                                                                                                                                                                                                                                                                                                                                                                                                                            Return the sum of the inputs. &                                                                                                                                                                                      Given the input numbers 9 and 7, the answer is 16.\textbackslash n\textbackslash n \\
                                               subtract\_two &      100 &                                                                                                                                                                                                                                                                                                                                                                                                                                                                                                                                                                                                                                                                                                                                                                                                                                                                                                                                                                                                     Return the difference of the inputs. &                                                                                                                                                                                       Given the input numbers 5 and 4, the answer is 1.\textbackslash n\textbackslash n \\
                                                 divide\_two &      100 &                                                                                                                                                                                                                                                                                                                                                                                                                                                                                                                                                                                                                                                                                                                                                                                                                                                                                                                                                                                                       Return the quotient of the inputs. &                                                                                                                                                                                     Given the input numbers 2 and 7, the answer is 2/7.\textbackslash n\textbackslash n \\
                                               multiply\_two &      100 &                                                                                                                                                                                                                                                                                                                                                                                                                                                                                                                                                                                                                                                                                                                                                                                                                                                                                                                                                                                                        Return the product of the inputs. &                                                                                                                                                                                       Given the input numbers 3 and 3, the answer is 9.\textbackslash n\textbackslash n \\
                                                    max\_two &      100 &                                                                                                                                                                                                                                                                                                                                                                                                                                                                                                                                                                                                                                                                                                                                                                                                                                                                                                                                                                                                        Return the maximum of the inputs. &                                                                                                                                                                                       Given the input numbers 1 and 1, the answer is 1.\textbackslash n\textbackslash n \\
                                   task1191\_food\_veg\_nonveg &      101 &                                                                                                                                                                                                                                                                                                                                                                                                                                                                                                                                                                                                                                                                                                                                                                                                                                                                                                                                                                            Return whether the input food dish is vegetarian (yes or no). &                                                                                                                                                                                                                Input: Haq Maas Answer: no\textbackslash n \\
                                 task1149\_item\_check\_edible &      119 &                                                                                                                                                                                                                                                                                                                                                                                                                                                                                                                                                                                                                                                                                                                                                                                                                                                                                                                                                                                     Return whether the input item is edible (yes or no). &                                                                                                                                                                                                                    Input: vase Answer: no\textbackslash n \\
                                   task1146\_country\_capital &      231 &                                                                                                                                                                                                                                                                                                                                                                                                                                                                                                                                                                                                                                                                                                                                                                                                                                                                                                                                  In this task, you are given a country name and you need to return the capital city of the given country &                                                                                                                                                                                     Input: Saint Pierre and Miquelon Answer: Saint-Pierre\textbackslash n \\
                                  task1147\_country\_currency &      232 &                                                                                                                                                                                                                                                                                                                                                                                                                                                                                                                                                                                                                                                                                                                                                                                                                                                                                                                                                   You are given a country name and you need to return the currency of the given country. &                                                                                                                                                                                                    Input: Senegal Answer: CFA Franc BCEAO\textbackslash n \\
                                task1509\_evalution\_antonyms &      551 &                                                                                                                                                                                                                                                                                                                                                                                                                                                                                                                                                                                                                                                                                                                                                                                                                                                                                             In this task, you are given an adjective, and your job is to generate its antonym. An antonym of a word is a word opposite in meaning to it. &                                                                                                                                                                                                              Input: paper Answer: scissor\textbackslash n \\
                                   task183\_rhyme\_generation &      999 &                                                                                                                                                                                                                                                                                                                                                                                                                                                                                                                                                                                                                                                                                                                                                                                                                                                                                                                           Given an input word generate a word that rhymes exactly with the input word. If not rhyme is found return "No" &                                                                                                                                                                                                                 Input: think Answer: sync\textbackslash n \\
                             task107\_splash\_question\_to\_sql &     2031 & In this task you are expected to write an SQL query that will return the data asked for in the question. An SQL query works by selecting data from a table where certain conditions apply. A table contains columns where every row in that table must have a value for each column. Every table has a primary key that uniquely identifies each row, usually an id. To choose which columns are returned you specify that after the "SELECT" statement. Next, you use a "FROM" statement to specify what tables you want to select the data from. When you specify a table you can rename it with the "AS" statement. You can reference that table by whatever name follows the "AS" statement. If you want to select data from multiple tables you need to use the "JOIN" statement. This will join the tables together by pairing a row in one table with every row in the other table (Cartesian Product). To limit the number of rows returned you should use the "ON" statement. This will only return rows where the condition... & Input: What are the order ids and customer ids for orders that have been Cancelled, sorted by their order dates? Answer: SELECT order\_id ,  customer\_id FROM customer\_orders WHERE order\_status\_code  =   "Cancelled" ORDER BY order\_date\textbackslash n \\
                         task088\_identify\_typo\_verification &     6499 &                                                                                                                                                                                                                                                                                                                                                                                                                                                                                                                                                 The given sentence contains a typo which could be one of the following four types: (1) swapped letters of a word e.g. 'niec' is a typo of the word 'nice'. (2) missing letter in a word e.g. 'nic' is a typo of the word 'nice'. (3) extra letter in a word e.g. 'nicce' is a typo of the word 'nice'. (4) replaced letter in a word e.g 'nicr' is a typo of the word 'nice'. You need to identify the typo in the given sentence. To do this, answer with the word containing the typo. &                                                                                                                                                                        Input: A laege display of apples, pears, and oranges Answer: laege\textbackslash n \\
task1336\_gender\_classifier &     6500 &                                                                                                                                                                                                                                                                                                                                                                                                                                                                                                                                                                                                                                                                                                                                                                                                                                                                                                                                                                                   Return the gender of the person in the input sentence. &                                                                                                                                                                                         Input: Justin made me feel discouraged. Answer: M\textbackslash n \\
                         task092\_check\_prime\_classification &     6500 &                                                                                                                                                                                                                                                                                                                                                                                                                                                                                                                                                                                                                                                                                                                                                                                                                                In this task, you need to output 'Yes' if the given number is a prime number otherwise output 'No'. A 'prime number' is a a whole number above 1 that can not be made by multiplying other whole numbers. &                                                                                                                                                                                                                   Input: 9319 Answer: Yes\textbackslash n \\
\bottomrule
\end{tabular}
    }
    \label{tab:data_details}
\end{table}

\begin{table}[H]
    \centering
    \small
    \caption{Models analyzed here.}
    \makebox[\textwidth][c]{
        \begin{tabular}{l c c}
    \toprule
     Model name & Huggingface identifier & Citation \\
     \midrule
     GPT-2 (1.5B) & \texttt{gpt2-xl} & ~\cite{radford2019language}\\
     OPT (2.7B) & \texttt{facebook/opt-2.7b} & ~\cite{zhang2022opt}\\
     GPT-Neo (2.7B) & \texttt{EleutherAI/gpt-neo-2.7B} & ~\cite{gpt_neo}\\
     Flan-T5 (3B) & \texttt{google/flan-t5-xl} & ~\cite{flan}\\
     GPT-J (6B) & \texttt{EleutherAI/gpt-j-6B} & ~\cite{gpt_j}\\
     OPT (6.7B) & \texttt{facebook/opt-6.7b} & ~\cite{zhang2022opt}\\
     Galactica (6.7B) & \texttt{facebook/galactica-6.7b} & ~\cite{taylor2022galactica}\\
     GPT-Neo (20B) & \texttt{EleutherAI/gpt-neox-20b} & ~\cite{black2022gpt}\\
     GPT-3 (175B) & \texttt{text-davinci-002} (OpenAI API) & ~\cite{radford2021learning}\\
    \bottomrule
\end{tabular}
    }
    \label{tab:models}
\end{table}
% \begin{table}[H]
%     \centering
%     \scriptsize
%     \caption{Template used for each dataset.}
%     \makebox[\textwidth][c]{
%     \input{tabs/data_queries}
%     }
%     \label{tab:data_queries}
% \end{table}

\subsection{\methods results extended}
\label{subsec:rank_exps_extended}

\begin{figure}[ht]
    \centering
    \includegraphics[width=0.5\textwidth]{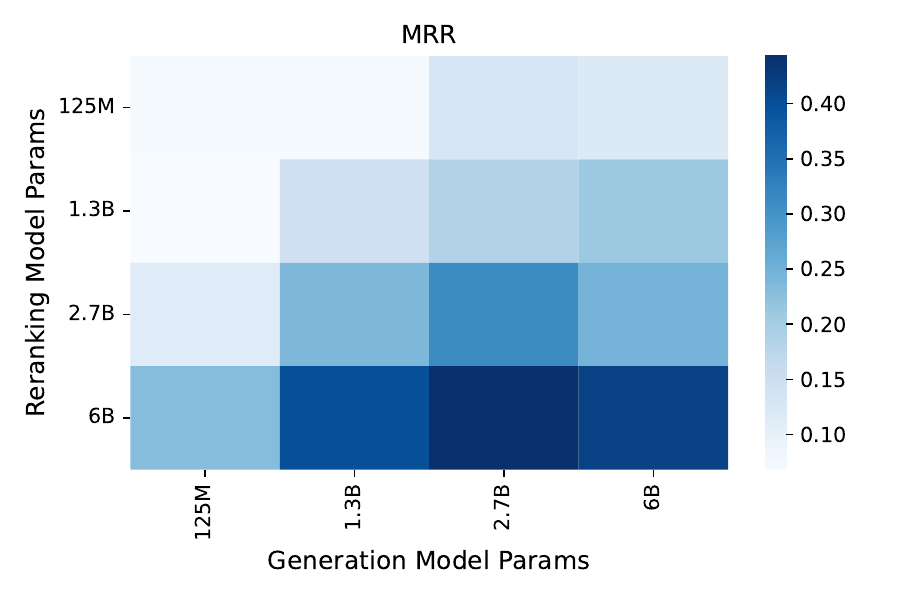}
    \vspace{-12pt}
    \caption{iPrompt performance across different size language models for the prompt proposal and reranking steps. Values are mean reciprocal rank of first accepted prompt averaged across 20 tasks and 3 random seeds.}
    \label{fig:ablation_disc_heatmap}
\end{figure}

We consider discriminators of varying sizes, with GPT-J (6B) as a prompt generator. We also compare generators of varying sizes with GPT-J (6B) as a prompt discriminator.
Models considered are of $\{125M, 1.3B, 2.7B, 6B\}$ parameters from the GPT-Neo/GPT-J language model family.
Results are shown in \cref{fig:ablation_disc_heatmap}. Performance varies smoothly across model sizes, with the highest performance when using the largest model for both reranking and generation. Reranking appears slightly more important than generation. When using a 1.3B parameter model for generation, MRR drops only slightly, from $0.418$ to $0.399$, while when using a 1.3B parameter model for reranking, MRR drops to $0.211$. In general, prompt recovery performance improves smoothly with reranking model size.

\begin{figure}[ht]
    \centering
    % \vspace{-25pt}
    \includegraphics[width=0.5\textwidth]{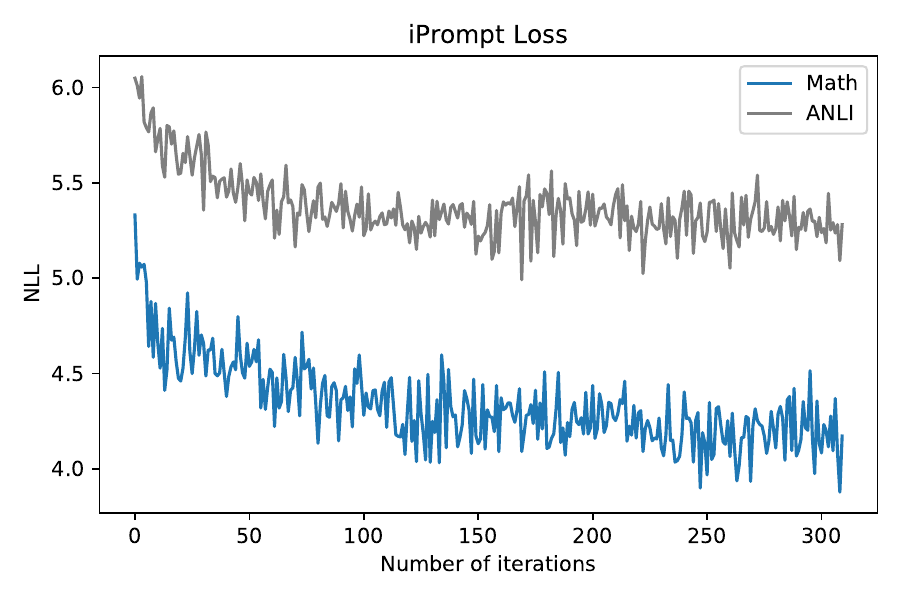}
    \vspace{-20pt}
    \caption{iPrompt performance across training, averaged across three random seeds and all tasks from Math datasets (Blue) and ANLI (Gray).}
    \label{fig:ablation_loss_convergence}
\end{figure}

\cref{fig:ablation_loss_convergence} plots the progress of iPrompt across iterations, comparing runs on Math datasets (blue) to runs on ANLI datasets (gray).
iPrompt appears to make most of its progress during the first $20\%$ of training and then continue to slowly decrease the average loss.
Running for more iterations on additional datapoints would likely increase performance.

\begin{table}[H]
    \centering
    \caption{Performance of Galactica at prompt recovery, including DD datasets~\cite{zhong2022describing,zhong2021adapting}.}
    \begin{tabular}{lllll}
\toprule
 & & iPrompt & AutoPrompt & Suffix \\
\midrule
\multirow[c]{4}{*}{MRR} & Math & \textbf{0.2} & 0.09 & 0.025 \\
 & ANLI & \textbf{0.39} & 0.0025 & 0.085 \\
 & Induction & \textbf{0.14} & 0.098 & 0.056 \\
 & DD & \textbf{0.064} & 0.0082 & 0.066 \\
\multirow[c]{4}{*}{Correct} & Math & \textbf{0.12} & 0.075 & 0 \\
 & ANLI & \textbf{0.34} & 0 & 0.025 \\
 & Induction & \textbf{0.071} & 0.087 & 0.02 \\
 & DD & \textbf{0.043} & 0 & 0.052 \\
\multirow[c]{4}{*}{BLEU-Top Prompt} & Math & \textbf{0.0073} & 0 & 0 \\
 & ANLI & \textbf{0.01} & 0 & 0.00032 \\
 & Induction & \textbf{0.022} & 0 & 0.0027 \\
 & DD & \textbf{0} & 0 & 0.0015 \\
\bottomrule
\end{tabular}
\end{table}
\FloatBarrier

\begin{table}[H]
    \centering
    \tiny
    {
    \renewcommand{\arraystretch}{1.4}
    \caption{Examples of top-generated prompts for each method: GPT-J main datasets.}
    \makebox[\textwidth][c]{
    \begin{tabular}{p{0.15\textwidth}p{0.35\textwidth} p{0.3\textwidth} p{0.25\textwidth}}
\toprule
 &                                                 autoprompt &                                               iprompt &                                                      suff \\

\midrule
active to passive                                           &                                \verb| (= 18 the the subst| &            \verb| Choose a pronoun for each sentence| &                     \verb| Create a sentence or group of| \\
add two                                                     &            \verb|>:Returns Adding togetherFont accomplish| &                  \verb| Create a function named `sum| &                                           \verb| n>2  m1| \\
antonyms                                                    &                                  \verb| the bectheBut But| &                    \verb| The noun to its opposite (| &                           \verb| The code to ascend. You| \\
cause and effect                                            &         \verb| REG Kinect virginity developed mosquit The| &                \verb| What would each sentence be if| &               \verb| write programs that read through an| \\
common concept                                              &       \verb|???????? parted configuredthe        ????????| &                 \verb| Find a noun that includes all| &                       \verb| which is a common word used| \\
diff                                                        &                             \verb|""Fair 62 disgust 92 81| &          \verb|  Find the difference between largest| &                    \verb| Write a program or function to| \\
divide two                                                  &         \verb| soughtWomen surgicalthe Percentage treated| &                         \verb| "Divide each digit by| &                   \verb| write a program or function who| \\
double one                                                  &               \verb| says transit Farethe doubles dollars| &               \verb| Write a function called double_| &                  \verb| Given two function pointer A and| \\
exp one                                                     &                          \verb|&&wl +# 123 270 Earthquake| &             \verb| Input this into your calculator (| &                       \verb| Type in number between 15 &| \\
fibonacci one                                               &                              \verb| baptpi produce347).''| &              \verb| Implement a function to find Fib| &                             \verb| Given an integer n (1| \\
first two                                                   &          \verb| Binding decode wr detect shortest numeric| &              \verb| Find first digit of given number| &                   \verb| When was Python added to Ubuntu| \\
first word letter                                           &                    \verb|Exception Ps< endoftext >the the| &                  \verb| Make a program that reads in| &                                        \verb| nimshul, a| \\
informal to formal                                          &     \verb|CLASSIFIEDthe themselves strongly Plays Chamber| &       \verb| These are questions on simple sentences| &  \verb| Make the following sentences positive statements| \\
larger animal                                               &                             \verb|????????thethethethethe| &                \verb| What is the most common animal| &                              \verb| dogAnswer to "What's| \\
letters list                                                &                              \verb| fluidsthethethethethe| &            \verb| Given the following list of tokens| &              \verb| The computer will make this document| \\
max two                                                     &                            \verb| spendingthethethethethe| &                  \verb| Implement a version of max()| &                      \verb| Write code to find out given| \\
multiply two                                                &           \verb|ruits="# multipl integer multiplied False| &                        \verb| 'How do you multiply a| &                   \verb| write a program or function who| \\
negation                                                    &                   \verb| performs antiv Sizethe NULL NULL| &             \verb| I found these four mistakes below| &                      \verb|  Your friends think that you| \\
num to verbal                                               &          \verb| irritatedthedd respectfully Protectivethe| &               \verb| Output each number below in the| &               \verb| The program outputs the first input| \\
orthography starts with                                     &                     \verb| nextbusiness wordevery morphpp| &                        \verb| Name of two homophones| &                      \verb| You will be given five words| \\
rhymes                                                      &           \verb| Steal batter dating: unfold testosterone| &                 \verb| Find the missing word for all| &                              \verb|  Input [create] What| \\
second word letter                                          &                                  \verb| i  mascot okay kk| &                        \verb| Who gave the answer "o| &            \verb|  the United states government outlawed| \\
sentence similarity                                         &                    \verb|                 value %%%% Math| &                              \verb| 3 (5 marks). The| &              \verb| Read five sentences about your topic| \\
sentiment                                                   &                 \verb| positively optimistic&&&& negative| &                         \verb| I'm voting "negative"| &                                   \verb| Melvins at CBGB| \\
singular to plural                                          &            \verb|Enhanced shorthand Lets pluralbetweenthe| &                   \verb| Given a noun and its plural| &                                     \verb| 1.  It may be| \\
square one                                                  &                          \verb|Cal impl qApplySquare fiat| &            \verb| Input number and return its square| &                    \verb| Write a program or function to| \\
subtract two                                                &                               \verb|ignorethethethethethe| &           \verb| Write a function to find difference| &                      \verb| Given a non-negative integer| \\
sum                                                         &                              \verb| Photosthethethethethe| &             \verb| Add two numbers together and then| &                  \verb| The program outputs, without any| \\
synonyms                                                    &                       \verb|Word termOn English meanings | &                         \verb| what is a synonym for| &                            \verb| Is there a cure for an| \\
task088 identify typo verification                          &                                   \verb|  thethethethethe| &                 \verb| This word scramble is to test| &                        \verb| You wake up in the morning| \\
task092 check prime classification                          &                         \verb|ropheospels&& Norestricted | &             \verb| Are these pairs of integers prime| &                  \verb| Print the input numbers in order| \\
task107 splash question to sql                              &                                                    \verb|| &                     \verb| How Do You Connect SQL To| &                       \verb| To get into MySQL you first| \\
task1146 country capital                                    &                  \verb| Ang Suppose AUTHthe beh Assassins| &                 \verb| Which city is the capital and| &                         \verb| France, England or the UK| \\
task1147 country currency                                   &                             \verb|aaaathecurrency Nib Sc | &             \verb| Ireland. Which currency is spoken| &                                \verb| "I am working on a| \\
task1149 item check edible                                  &                                   \verb| no the870830 yes| &                  \verb| coffee and beans are fruits.| &                     \verb| Which one of the following is| \\
task1191 food veg nonveg                                    &  \verb| compliedthe whether methamphetamine provided comp| &                        \verb| Are you a vegetarian? | &                             \verb| It could be any food,| \\
task1336 peixian equity evaluation corpus gender classifier &               \verb|< endoftext > -> M Fundamental FG Fav| &                       \verb| Predict the gender (F =| &                                        \verb|??????,???,| \\
task1509 evalution antonyms                                 &          \verb| contrad orously inverted ironically trans| &          \verb| find words with the opposite meaning| &                     \verb| Record your input and answer,| \\
task183 rhyme generation                                    &                        \verb| quarterdream dug}. Thro rhy| &                           \verb| Mind vs Glee! There| &                           \verb| what do you love to eat| \\
taxonomy animal                                             &           \verb| programmingQ errorsBefore admitting mont| &              \verb| What are the most common animals| &                      \verb| Each of these questions is a| \\
translation en-de                                           &                               \verb| H prob Hyper  Forthe| &                \verb| You are a lawyer practicing in| &                       \verb| This is an example of input| \\
translation en-es                                           &                                \verb|  the ththebb volunt| &       \verb| please write English meaning in Spanish| &                                          \verb|  Porque?| \\
translation en-fr                                           &                       \verb| IRthe< endoftext >thethe the| &                   \verb| What is the French word for| &                      \verb| Your code needs to deal with| \\
word in context                                             &                        \verb| ("nSame distinguishedthethe| &                           \verb| Same and Not-Same -| &                          \verb| What you will do is have| \\
\bottomrule
\end{tabular}

    }
    }
    \label{tab:prompt_examples_full}
\end{table}

\begin{table}[H]
    \centering
    \tiny
    {
    \renewcommand{\arraystretch}{1.4}
    \caption{Examples of top-generated prompts for each method: GPT-J DD datasets~\cite{zhong2022describing,zhong2021adapting}.}
    \makebox[\textwidth][c]{
    \begin{tabular}{p{0.05\textwidth}p{0.35\textwidth} p{0.3\textwidth} p{0.25\textwidth}}
\toprule
 &                                                 autoprompt &                                               iprompt &                                                      suff \\
                                                   &                                                            &                                                       &                                                           \\
\midrule
d3 0                                                        &                                                    \verb|| &                 \verb| line contains this string? No| &                          \verb| contains all 6 items, No| \\
d3 1                                                        &  \verb| Ghostbustersthe interrogation condition criminall| &                    \verb| sentence contains "yes" or| &                 \verb| string doesn't match any template| \\
d3 10                                                       &          \verb| preceded Roosevelt nonexistentuphem_-_ Tw| &                     \verb| message contains "no". No| &                    \verb| contains all of these words or| \\
d3 11                                                       &           \verb| caused senator  prompt Recall interacted| &                       \verb| string contains "No" or| &                  \verb| was matched; output otherwise No| \\
d3 12                                                       &                        \verb| begin:" r "},{" contradict| &                          \verb| tweet mentions  yes | &                        \verb| is true or output false if| \\
d3 13                                                       &                                 \verb|},{"    vote [*"]=>| &                               \verb| answer "no" (or| &                  \verb| contains all correct answers, No| \\
d3 14                                                       &   \verb| nonexistent undead questions Enhance mandated no| &                        \verb| string begins 'no' and| &               \verb| string contains any non blank white| \\
d3 15                                                       &              \verb| rarely ----Question not},{" geometric| &                       \verb| string contains "no" or| &           \verb| includes exactly two English words with| \\
d3 16                                                       &                           \verb|\n pearthemar Display RUN| &                      \verb| text contains any "yes".| &                     \verb| text is true, otherwise write| \\
d3 17                                                       &                        \verb| EMP Similarly\t=== charsthe| &                           \verb| is an answer ("no",| &             \verb| contains all correct answers for this| \\
d3 18                                                       &                         \verb|\n\n Verb horm  suffix Eucl| &                      \verb| phrase starts with 'no',| &              \verb| contains all correct answers else No| \\
d3 19                                                       &                   \verb|\n."," Emacs strips colors strips| &                       \verb| word starts with 'yes',| &                \verb| text contains any of these strings| \\
d3 2                                                        &        \verb| indirectly [[ pervasive?"Spoiler exhaustive| &                           \verb| ends with "yes". If| &                               \verb| sentence has an "O"| \\
d3 20                                                       &                        \verb|\n\n dips Vote flower Ainthe| &                \verb|ted sentence contains both "yes| &                    \verb| contains one of these words or| \\
d3 21                                                       &                        \verb|\nthePubLeft        Abstract| &                            \verb| ends with 'no'. No| &                  \verb| contains all correct answers, or| \\
d3 22                                                       &                              \verb|Nov wholesno Eucl NO  | &                            \verb| can output no/yes,| &              \verb| data set contains results for output| \\
d3 23                                                       &       \verb| vantage immediately recogn example nails 309| &                    \verb| no else output none? Input| &          \verb| contains data describing or referring to| \\
d3 24                                                       &                                   \verb| noBER nonosRew [| &            \verb| datum defines finite number fields| &                               \verb| is in fact equal 2;| \\
d3 25                                                       &                    \verb| withdrawalsnob  inher nob Among| &             \verb| contains both gene list data file| &                    \verb| has already started in state x| \\
d3 26                                                       &               \verb|Joined robberHigthe contradictionNarr| &                       \verb| line ends with a space,| &                    \verb|ted series matches any of these| \\
d3 27                                                       &               \verb| verseoleon:- inferred   cannabinoids| &                   \verb| was positive answer and "No| &                         \verb| string of words, as shown| \\
d3 28                                                       &                                \verb|\n repet999 REM=[nov| &         \verb| refers exclusively (only literally or| &                    \verb| was a real question that could| \\
d3 29                                                       &             \verb|\n Pat uncertaintiesMerit         oppos| &                         \verb| line begins with  yes| &                        \verb| text meets any one or more| \\
d3 3                                                        &                              \verb|\n\n887odynamHor mun\t| &                           \verb| ends with "yes" and| &        \verb| statement reflects truth. Otherwise output| \\
d3 30                                                       &                       \verb| detainees gap   ${. hardness| &                \verb| statement is false?  Otherwise| &                  \verb| is an example from each category| \\
d3 31                                                       &                    \verb|\n055 helium        **** itching| &             \verb| phrase does not contain any words| &                          \verb| given was false or not a| \\
d3 32                                                       &                                 \verb|      Afghthethethe| &           \verb| matches either one of these strings| &                           \verb| text is true, and write| \\
d3 33                                                       &                                    \verb| le    \r 253   | &                 \verb| has a duplicate word. Correct| &                                    \verb| contains  yes | \\
d3 34                                                       &                         \verb| the Carnegie allerg Qu the| &                                  \verb| no,no for (1| &                                  \verb| was "The End" or| \\
d3 35                                                       &                         \verb| Hatch Land pri poker[[ Yah| &                              \verb| would be a no (I| &                   \verb| text can create a good argument| \\
d3 36                                                       &                                \verb|],  egregbyte?Sensor| &                       \verb| matches exactly a "no".| &                      \verb| string meets any, or exactly| \\
d3 37                                                       &                         \verb| noun441...? word first neg| &                    \verb| question has an answer "no| &                       \verb| string meets any, and write| \\
d3 38                                                       &                  \verb| wond <+ HELP"},{"InvalidOtherwise| &                                    \verb| says  yes | &                                     \verb| "yes"  has an| \\
d3 39                                                       &                                  \verb| notnobbutthe but | &                               \verb| reads like  no.| &                             \verb| answers "yes" for all| \\
d3 4                                                        &                 \verb|\n\n 760 consensualNarr Fog cabbage| &                      \verb| sentence ends with "no".| &               \verb| string was a valid answer otherwise| \\
d3 40                                                       &                                    \verb|modeXP/, \n  but| &               \verb| question contains an actual "no| &                   \verb| given was wrong or not relevant| \\
d3 41                                                       &        \verb| opinions universitythe began followingawaru| &            \verb| sentence is grammatically correct,| &                                \verb| equals to zero (i.| \\
d3 42                                                       &  \verb| disqualified hemor Ratings [  contradiction Moham| &  \verb| phrase represents something that is actually| &                          \verb| has 1 out of 2 responses| \\
d3 43                                                       &                    \verb|\n\n saturated         Phot misc| &                      \verb| would be rightAnswer :no| &               \verb| was about a government regulation (| \\
d3 44                                                       &                                  \verb|\n  <[ npm spaces1| &                              \verb| was "no":  Input| &                              \verb| was "yes" else false| \\
d3 45                                                       &                             \verb|\n\n pit  VerbFalse Tok| &                     \verb| string contains one "no".| &                            \verb| text starts with "OK",| \\
d3 46                                                       &                                 \verb|  },{" Neil kingthe| &               \verb| no when a string containing one| &                        \verb| contains this string! Yes,| \\
d3 47                                                       &                      \verb| network intuitive     19 Lamp| &             \verb| sentence implies that no can mean| &                      \verb| contains all digits, else No| \\
d3 48                                                       &               \verb| nond307 Literally negativeJun corpor| &                  \verb| conforms with known facts no| &                        \verb|ted number from user base 5| \\
d3 49                                                       &                               \verb|Falsethe    Rect 802| &                       \verb| string contains "no" or| &                      \verb| contains all of these words,| \\
d3 5                                                        &            \verb| contradicts absurdity Luffythe neg answ| &                        \verb| string 'no' appears as| &                        \verb| is correct  ; No otherwise| \\
d3 50                                                       &          \verb|  ________________________  WithNo","hedon| &                             \verb| mentions "no" (or| &                    \verb| contains all correct items, No| \\
d3 51                                                       &                 \verb|\n\n 276WithNo   noodles Cosponsors| &                            \verb| reads "no" no else| &                          \verb| given was no; not output| \\
d3 52                                                       &                              \verb|\n\n 225Should  laure | &                           \verb| string was 'no' and| &                   \verb| string contains just one space.| \\
d3 53                                                       &                               \verb|never_{ Johns  neo no| &                    \verb| is all lower case answer 1| &                       \verb| was what I described above!| \\
d3 6                                                        &       \verb| forbids         Literally reminisNone negate| &                        \verb| text contains any "no"| &                            \verb| text contains  Syrian | \\
d3 7                                                        &                      \verb|},{"\r stringologically $\ git| &                       \verb| contains 'no' or output| &                               \verb| text contains  yes | \\
d3 8                                                        &      \verb| unlikelyEitherselessletter Ches contradictory| &                     \verb| sentence contains 'no' or| &              \verb| contains any newlines after matching| \\
d3 9                                                        &                     \verb| reactive happensMiddle lot Inc| &                       \verb| matches any word (no is| &                           \verb| text meets any, or none| \\
\bottomrule
\end{tabular}

    }
    }
    \label{tab:prompt_examples_full_dd_gptj}
\end{table}

\begin{table}[H]
    \centering
    \tiny
    {
    \renewcommand{\arraystretch}{1.4}
    \caption{Examples of top-generated prompts for each method: Galactica main datasets.}
    \makebox[\textwidth][c]{
    \begin{tabular}{p{0.15\textwidth}p{0.35\textwidth} p{0.3\textwidth} p{0.25\textwidth}}
\toprule
 &                                                 autoprompt &                                              iprompt &                                                    suff \\
                                                   &                                                            &                                                      &                                                         \\
\midrule
active to passive                                           &  \verb| Transmission Electthe chromosome initialized empl| &                        \verb| 4-way Multiple Choice| &                  \verb| Is the context a good response| \\
add two                                                     &                                \verb|  addthe Hyper  addi| &                       \verb| In order to add two or| &                  \verb| Given three real-valued inputs| \\
antonyms                                                    &           \verb| meet equilibration stiptertead asymmetry| &                 \verb| What is the opposite of each| &                                   \verb| [T1] Question| \\
cause and effect                                            &                              \verb| shaking Dthethethethe| &                    \verb| Find clues as to why each| &                   \verb| What do you think will happen| \\
common concept                                              &                             \verb| Bary techntbltbltbl Te| &                   \verb| Where are all the animals?| &                         \verb| What' s the most common| \\
diff                                                        &      \verb| quartic digits shorter recreational genomics| &            \verb| Given two positive integers a and| &                       \verb| What's the most efficient| \\
divide two                                                  &      \verb| manipulations comput  iterationects quotients| &                     \verb| The ratio of two real or| &      \verb| Given two different positive integers what| \\
double one                                                  &                        \verb| roll Add Pingthe brakingthe| &              \verb| Determine how much money did Al| &                               \verb| What's it like to| \\
exp one                                                     &                        \verb| visc poplLSPLC Viscositythe| &                      \verb| Given a number y and an| &                  \verb| Find a formula for this linear| \\
fibonacci one                                               &           \verb| start Attstrass Prim Polynomial emotions| &                                \verb|  \bigcirc m o| &                  \verb| Write a function that gives an| \\
first two                                                   &                                 \verb|AICthethethe Adethe| &      \verb| Solve using negative exponents? Explain| &                     \verb| We have found it helpful to| \\
first word letter                                           &                                \verb| d rthe l c syllable| &                       \verb| What is the last word?| &                                 \verb| the program {x.| \\
informal to formal                                          &        \verb| Why unpredictable comprobablyould Detecting| &                       \verb| Yes! However, since we| &                               \verb| Text-to-Text Data| \\
larger animal                                               &                      \verb| sharkoganopeanionaller descri| &              \verb| A question is given about three| &                       \verb| Is the pair of animals on| \\
letters list                                                &                  \verb| microm phon te photothermal te te| &                      \verb| How many 8 letter words| &         \verb| Given the following paragraph, indicate| \\
max two                                                     &                          \verb|$$amater Penet  credible b| &                   \verb| How large was each of your| &                    \verb| Is that as simple or complex| \\
multiply two                                                &        \verb|aris visualthe Gibson multiplicative lexical| &                 \verb| When we multiply two even or| &               \verb| What number divided by what other| \\
negation                                                    &                   \verb| brood he Apparent denselythe FIG| &                \verb| What did these people have as| &                      \verb| This time we do two prompt| \\
num to verbal                                               &    \verb| Pixel lum sedimentary precedenceathion thousand| &                               \verb| P(data answer)| &                    \verb| Number pairs that are in the| \\
orthography starts with                                     &  \verb|criptions geochemistry Harvey preprocessed Kus Cap| &            \verb| The correct verb after each input| &               \verb| Why did they choose this strategy| \\
rhymes                                                      &    \verb| hallucinations song cooperationcorner ask smear| &                        \verb| Which phrase did "sea| &                          \verb| My favorite food is a | \\
second word letter                                          &                              \verb|oderraj dialectath u o| &                   \verb|  What is the fourth letter| &                     \verb| Is the object in this image| \\
sentence similarity                                         &                   \verb|false provleastleast   Apparently| &  \verb| I understand your definition correctly that| &                        \verb| Chinese No Vote and Euro| \\
sentiment                                                   &         \verb| nominationnegative<unk>indolinivalentpolar| &                   \verb| What is the sentiment of a| &                   \verb| What do you think will happen| \\
singular to plural                                          &                               \verb| mes sequthethethethe| &                   \verb| Find the pluralization of | &                       \verb| Do you have any good ways| \\
square one                                                  &            \verb| AnalyticmassesAtomnamespace binning pow| &              \verb| Determine how much money did Al| &                               \verb| What's it like to| \\
subtract two                                                &                         \verb|ComplexRemthe scienti Event| &              \verb| Given a variable called A whose| &                    \verb| Is that close to your actual| \\
sum                                                         &                                \verb| Horujanthethethethe| &                          \verb| I'm trying to solve| &                   \verb| Is the following number even?| \\
synonyms                                                    &  \verb| straightforward conceptual Striking Etymology tra| &                      \verb| Can you think of a word| &                                           \verb| [T1],| \\
task088 identify typo verification                          &               \verb| Etymology nom scalesrolateral QMples| &                     \verb| What is the plural form?| &                  \verb| Other types Task Definition ::| \\
task092 check prime classification                          &                  \verb| Accept No source   Inter question| &                               \verb| Q3_NoAnswerYes| &                 \verb| Are there any types of chemical| \\
task107 splash question to sql                              &                                                    \verb|| &               \verb| Question answering Input #Name| &          \verb| Is the following SQL clause equivalent| \\
task1146 country capital                                    &                 \verb| Outer Hassan wal Tu Spontaneous Qu| &              \verb| List the capital cities in each| &                          \verb| The country that _____| \\
task1147 country currency                                   &                                    \verb| Llthethestr the| &             \verb| Find the most common currency in| &                  \verb| What currency was the first to| \\
task1149 item check edible                                  &                     \verb| nonthethe Characterizing Nothe| &                           \verb| Why is  no answer | &                                \verb| True or False, "| \\
task1191 food veg nonveg                                    &                      \verb|gue axiomsepid Output yes Birk| &                  \verb| Are you a native speaker of| &                    \verb| In a world where the Supreme| \\
task1336 peixian equity evaluation corpus gender classifier &                        \verb| lineage Mthe knockdown Fthe| &                       \verb| What is the gender of | &            \verb| Who is a good conversational partner| \\
task1509 evalution antonyms                                 &  \verb| Modern Carlson Weyl Linguistic counterfactual met| &              \verb| Find the opposite of each given| &                     \verb| We can predict text from an| \\
task183 rhyme generation                                    &                         \verb| stellarthethethe pl battle| &                            \verb| The 6-letter word| &               \verb| We are given a dataset consisting| \\
taxonomy animal                                             &                 \verb| duoull Pap codebook varic lysozyme| &            \verb| When two objects collide and expl| &                          \verb| What's the most common| \\
translation en-de                                           &                       \verb| shor Thanthe condens Intinte| &              \verb| Test for spelling error in word| &                  \verb| Is the object of your activity| \\
translation en-es                                           &               \verb| trophic Description params oscthethe| &                    \verb| In Spanish, there are two| &                              \verb| cuatro con la frec| \\
translation en-fr                                           &                              \verb| TT tic tgtthethe Disk| &                          \verb| Les champs du monde| &                      \verb| What can the words in bold| \\
word in context                                             &                               \verb|" Tang samethe offOff| &      \verb| Identify similar phrases based on given| &                 \verb| Does this sentence come from an| \\
\bottomrule
\end{tabular}

    }
    }
    \label{tab:prompt_examples_full_gal_main}
\end{table}

\begin{table}[H]
    \centering
    \tiny
    {
    \renewcommand{\arraystretch}{1.4}
    \caption{Examples of top-generated prompts for each method: Galactica DD datasets~\cite{zhong2022describing,zhong2021adapting}.}
    \makebox[\textwidth][c]{
    \begin{tabular}{p{0.15\textwidth}p{0.35\textwidth} p{0.3\textwidth} p{0.25\textwidth}}
\toprule
 &                                                 autoprompt &                                              iprompt &                                                    suff \\
                                                   &                                                            &                                                      &                                                         \\
\midrule
d3 0                                                        &                         \verb| Alloy ReeABL vetotitledthe| &     \verb| satisfies sarcastic predicate; otherwise| &                  \verb| is sarcastic, otherwise ignore| \\
d3 1                                                        &           \verb| Cosm compositionallyind locom astro bfnm| &                          \verb| and output share 82| &             \verb| sentence describes or is related to| \\
d3 10                                                       &                          \verb|onso Seman   NichentiVALID| &          \verb| paragraph does not contain any word| &                       \verb| says the answer is yes on| \\
d3 11                                                       &         \verb|enzo conspicuous Widespreadfeature cis orth| &                 \verb| mention e does not match any| &           \verb| says that the United States president| \\
d3 12                                                       &                   \verb|assert unco Nog antich DesignsFOR| &     \verb| contained a negation phrase otherwise an| &                 \verb| says that someone arrives or de| \\
d3 13                                                       &                       \verb| functionnoAns medi monos BAA| &           \verb| text contains no keywords and none| &                         \verb| is valid, no otherwise.| \\
d3 14                                                       &                                \verb|E PotassiumztheANASS| &         \verb|  the United Nations integrated multi| &           \verb| contains the context word or response| \\
d3 15                                                       &          \verb|no Nons TRANS Trajectories Exclusionifying| &                        \verb| phrase is not a noun;| &          \verb| example satisfies all rules, otherwise| \\
d3 16                                                       &                            \verb|TiHas Gomes immigPropthe| &               \verb| sentence contains the word  no| &                     \verb| mentions the answer and @US| \\
d3 17                                                       &  \verb| spatiotemporal extragalactic conflicts forbidden | &               \verb| data includes at least one Sem| &                      \verb| was true, and output false| \\
d3 18                                                       &               \verb| formulAns revisit   transcri neither| &                              \verb| ends in  no  no| &                       \verb| contain any formals in it| \\
d3 19                                                       &                     \verb| fatSPR Inhibitsickel nestedyes| &                          \verb| is valid.Answer: no| &                        \verb| text contains the word "| \\
d3 2                                                        &        \verb| propositional ScalarAsp Attacks train Rabin| &                                              \verb|| &            \verb| contain any of given words otherwise| \\
d3 20                                                       &                 \verb|Sem adjunct DCT Eriks admissibleArg| &                   \verb| is prochoice no otherwise | &          \verb| says something about abortion or human| \\
d3 21                                                       &                          \verb| scatterflows vetoriz  pen| &                   \verb| sentences contain both "no| &       \verb| sentence includes sexual, gender identity| \\
d3 22                                                       &               \verb|yesoscopyGal martingale Yes epistemic| &                                \verb| no. For ``yes| &    \verb| data satisfy certain conditions Otherwise No| \\
d3 23                                                       &                        \verb| NoELO predictors SBATCHvect| &    \verb| holds no otherwise [START_REF] Primordial| &          \verb| Predictive Models are Interpretable on| \\
d3 24                                                       &         \verb| norist Investigating Nos tumorigenesis Bit| &                           \verb| term "noisy inputs| &  \verb| follows the given probability density function| \\
d3 25                                                       &                   \verb| nopins bil field ensembles Locus| &                        \verb| no output no yea Prom| &          \verb| says that neutrinos have been observed| \\
d3 26                                                       &                                  \verb|NeuthePreftheDEthe| &                   \verb| sentence is a negation; an| &                       \verb| sentence includes "cutter| \\
d3 27                                                       &               \verb| no   Conditional abstract definiteLD| &            \verb| statement contains this word, and| &          \verb| says that certain events have happened| \\
d3 28                                                       &                       \verb|CIS raftriendrolimussubseteq | &                  \verb| data contains feminism, and| &                         \verb| says that are feminists| \\
d3 29                                                       &          \verb| noAns Semantic neitherHamiltonian dissoci| &                          \verb| text contains  no, | &          \verb| says something against women or gender| \\
d3 3                                                        &             \verb| nondec yes Census Tam Policies acyclic| &                              \verb| IS semst; else,| &         \verb| says something against your religion on| \\
d3 30                                                       &                     \verb|itasenta Assim allergic Fraser | &                 \verb| text contains answer=yes and| &                          \verb| data includes y and n,| \\
d3 31                                                       &            \verb| Strategy monitors Confl HaleFIELD Rhode| &          \verb| data contains a negative sentiment,| &                       \verb| matches at least one of a| \\
d3 32                                                       &        \verb| Regulates term Cliff steer VER Saskatchewan| &                         \verb| mentions  no  and no| &         \verb| sentence includes a pronoun that refers| \\
d3 33                                                       &                         \verb| mut Congress SyntN weakhis| &                \verb| text contains the phrase  yes| &              \verb| sentence includes a token for each| \\
d3 34                                                       &             \verb|yes<fragments> Kohn povertyyes Circular| &                      \verb| are based in movies. no| &                          \verb| says that Erik has his| \\
d3 35                                                       &                            \verb|noon nonlocalakh no no s| &           \verb| question contains YesNo words like| &              \verb| movie was very good otherwise mark| \\
d3 36                                                       &                     \verb| describes nomoduleno RevealsAs| &           \verb| sentence does not contain a factor| &                  \verb| text includes any unanswerable| \\
d3 37                                                       &                              \verb|penADOapineg autoclHAL| &                     \verb| phrase  no  appears only| &               \verb| sentence has an answer. Otherwise| \\
d3 38                                                       &                 \verb|noNoEnabl complementation BIT Polar| &             \verb| question contains the phrase no,| &             \verb| says that certain language has more| \\
d3 39                                                       &                    \verb| Neuastro neur runaway suffixthe| &           \verb| utterance contains this phrase  no| &          \verb| says something about your personality,| \\
d3 4                                                        &              \verb| MULT semilinear unarybuffer Gior fate| &            \verb| sentence does not contain a modal| &                \verb| meets any condition given in Sem| \\
d3 40                                                       &   \verb| outputs vigilance mK Unsupervised Status initial| &                 \verb| data contains no and no else| &      \verb| correctly answers your question, otherwise| \\
d3 41                                                       &             \verb|answ neph Membership Bess decomp neurop| &                \verb| equilibrium does not hold; no| &                    \verb| does not contain either of x| \\
d3 42                                                       &       \verb| Surveillance Semantics Obl Inhibits Hels MEL| &                      \verb| string isn't in English| &          \verb| says that climate issues have worsened| \\
d3 43                                                       &                                \verb|Ans yesArg Zika spar| &        \verb| supports my belief no otherwise Input| &           \verb| follows the context; Otherwise output| \\
d3 44                                                       &                    \verb|wer: inducible affirm Abl reflex| &                                              \verb|| &                    \verb| contain any formals words or| \\
d3 45                                                       &                        \verb| ana1 ERGsentence loopsyless| &            \verb| string does not occur in training| &                         \verb| question were "Is there| \\
d3 46                                                       &          \verb| GitHub Clevelandck negation RCC Microbial| &                     \verb| contains no fake or misn| &                 \verb| movie was released before year | \\
d3 47                                                       &                               \verb|ful eth massoc bis NA| &               \verb| debris affects doesnt have any| &               \verb| says that we need your assistance| \\
d3 48                                                       &                     \verb|\n Nons FernclassGridUHFFFAOYSA| &             \verb| holds for all possible inputs no| &             \verb| sentence includes a pronoun as well| \\
d3 49                                                       &                           \verb|noNo Imper Creating noPan| &                    \verb| sentence contains  no  in| &          \verb| matches answer which will give correct| \\
d3 5                                                        &           \verb| volat Salv Artificial economies fut Hale| &                    \verb| prompt is followed by  no| &                       \verb| says that the output is a| \\
d3 50                                                       &                             \verb|failedkin      ResDesMM| &             \verb| string does not contain any stop| &                   \verb| says that wight is decreasing| \\
d3 51                                                       &                  \verb| bl Frederthe Novo phylogeneticthe| &                             \verb| for "is my child| &           \verb| contains the context of your response| \\
d3 52                                                       &                           \verb|onasnono domainsex Quanti| &                     \verb| phrase has the value no,| &      \verb| sentence includes something that will lead| \\
d3 53                                                       &                                     \verb|onisenony anonh| &            \verb| includes the words no output will| &              \verb| contains at least two noun phrases| \\
d3 6                                                        &                    \verb| Alle substrthe Edmund Hos forks| &              \verb| answer no contains this word or| &                    \verb| is a valid response and vice| \\
d3 7                                                        &                              \verb|Antithethethe Blakethe| &                  \verb| word is a negation of micro| &     \verb| sentence includes all possible answers Prom| \\
d3 8                                                        &                \verb| Brand abolished  affili attri Recon| &         \verb| corresponds with prompt question  no| &           \verb| sentence is suitable Question for yes| \\
d3 9                                                        &                   \verb| Bou counterex abstnougin literal| &               \verb| question has answer no, output| &               \verb| is correct but maybe not relevant| \\
\bottomrule
\end{tabular}

    }
    }
    \label{tab:prompt_examples_full_gal_dd}
\end{table}

\subsection{Experiment details / hyperparameters extended}
\label{sec:exp_details_supp}

\begin{figure}[H]
    \centering
    \includegraphics[width=0.75\textwidth]{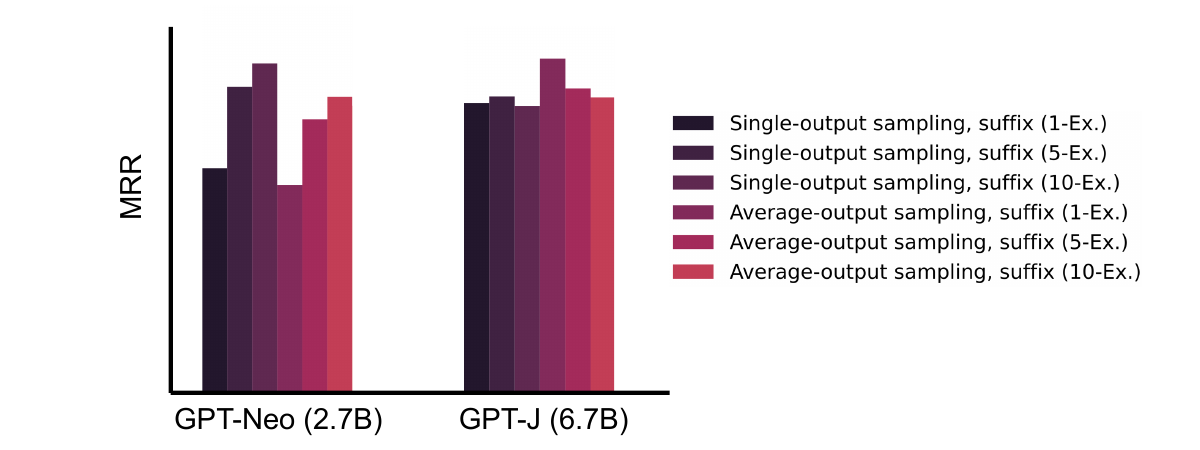}
    \vspace{-10pt}
    \caption{Average suffix sampling versus individual-example suffix sampling does not improve performance (for ANLI datasets).}
    \label{fig:avgsuffix_ex}
\end{figure}

\paragraph{Average-output suffix decoding}
LLMs themselves can be directly used to predict prompt strings.
We can give the model a prompt that includes examples such as the following context string:
$\underbrace{\textit{In: 2 5}}_{x^i} \underbrace{\textit{Out: 7.}}_{y^i} \underbrace{\textit{To compute the output from the input, }}_{\textit{template}}$\blank, and sample the output for the blank to recover a prompt $\hat s$.
Sampling directly from $f$ helps ensure that the generated explanation is fluent and semantically meaningful.
We decode the output using beam search to find the highest-probability outputs for multi-token prompts.\footnote{Here we prefer beam search here over alternatives such as nucleus sampling~\citep{holtzman2019curious} as we are interested in finding an accurate prompt description with as few samples as possible.}
To improve on this approach, we place several examples into the model's context, and then average the model's output logits across all the examples in the dataset before decoding the output, an approach we refer to as \textit{average-suffix decoding}.
However, we find that average-suffix decoding does not yield a performance improvement over straightforward decoding from a single sample with examples in the context.
For example, \cref{fig:avgsuffix_ex} shows that for the ANLI datasets, the mean reciprocal rank for average-output sampling does not tend to be higher than for single-output sampling across two
 different models.

\paragraph{Hyperparameters for iPrompt and AutoPrompt}
This subsection discusses the hyperparameters set for prompts generated on Math, NLI, and sentiment tasks. For Math and NLI tasks we considered prompts of length $6$ tokens; for sentiment we considered prompts of length $16$. For all experiments with \methods we consider $8$ candidate explanations for each step and generate $4$ new generations per candidate, for a total of $32$ candidates. For fair comparison, we consider $32$ candidates per step for AutoPrompt. We generate Math and NLI from $5,000$ training steps and Sentiment candidates from $10,000$ steps. We truncate examples to a maximum of $128$ tokens. We measure loss for re-ranking (used by both AutoPrompt and \method) using the LLM's loss over the full space of output tokens, i.e. we do not restrict the vocabulary to the space of label tokens for classification problems.

\paragraph{Details of \method}

Here we explicate the details of \method. At each step, we consider a fixed number of mutations for each example in the population, as well as an additional number of random generations to prevent the population from getting stuck in a local minimum. When we sample a new population, we sample the best-performing prompts seen so far, as measured by a running average zero-shot loss. In order to encourage diverse candidate prompts, sample a population such that each sample starts with a different token. During preliminary experiments, we found that enforcing different starting tokens for each candidate prompt helped promote more diverse and interpretable prefixes.

For generation, we sample directly from the LLM given the data concatenated with the string \texttt{\\nPrompt:}. We sample with a temperature of $1$ and do not use a sampling strategy like nucleus sampling. For Math and NLI, we set the ``repetition penalty'' for generations to $2.0$ to discourage copying from the training set. For the sentiment experiment, we reduce the repetition penalty to $1.0$.

\paragraph{Details of AutoPrompt}

We note several changes to AutoPrompt that were not mentioned in the original paper but present in the original codebase, and proved crucial in our implementation. 

First, if we compute the top-candidates over every position, the magnitude of the gradient will always be highest at position 0, and thus AutoPrompt will prefer to make a swap at that position every time. To fix this issue, at each training step, we randomly select a position of the token to edit and consider word swaps only at that position. 

Second, as described, AutoPrompt will always take one of the candidate substitutions, even when said candidate does not improve the loss compared to the current prefix. Instead, we only make a substitution if the candidate prefix loss is lower than the loss on the same batch computed with the current prefix. 

Finally, \textit{unlike} the AutoPrompt implementation found online, we allow AutoPrompt to select from any token to substitute, including special tokens and non-English characters. 

To make AutoPrompt compatible with ranking-based metrics, we store the losses for each candidate ranked during training. At the end, we consider the ``top prefix'' to be the prefix with the lowest average loss during training, that has been considered at least three times. This final consideration criteria prevents candidates from the very end of training that only have a few loss estimates from being counted as the top prefix.

\subsection{Galactica experiment details}
\label{sec:galactica_supp}

\begin{figure}[H]
    \centering
    \includegraphics[width=0.6\textwidth]{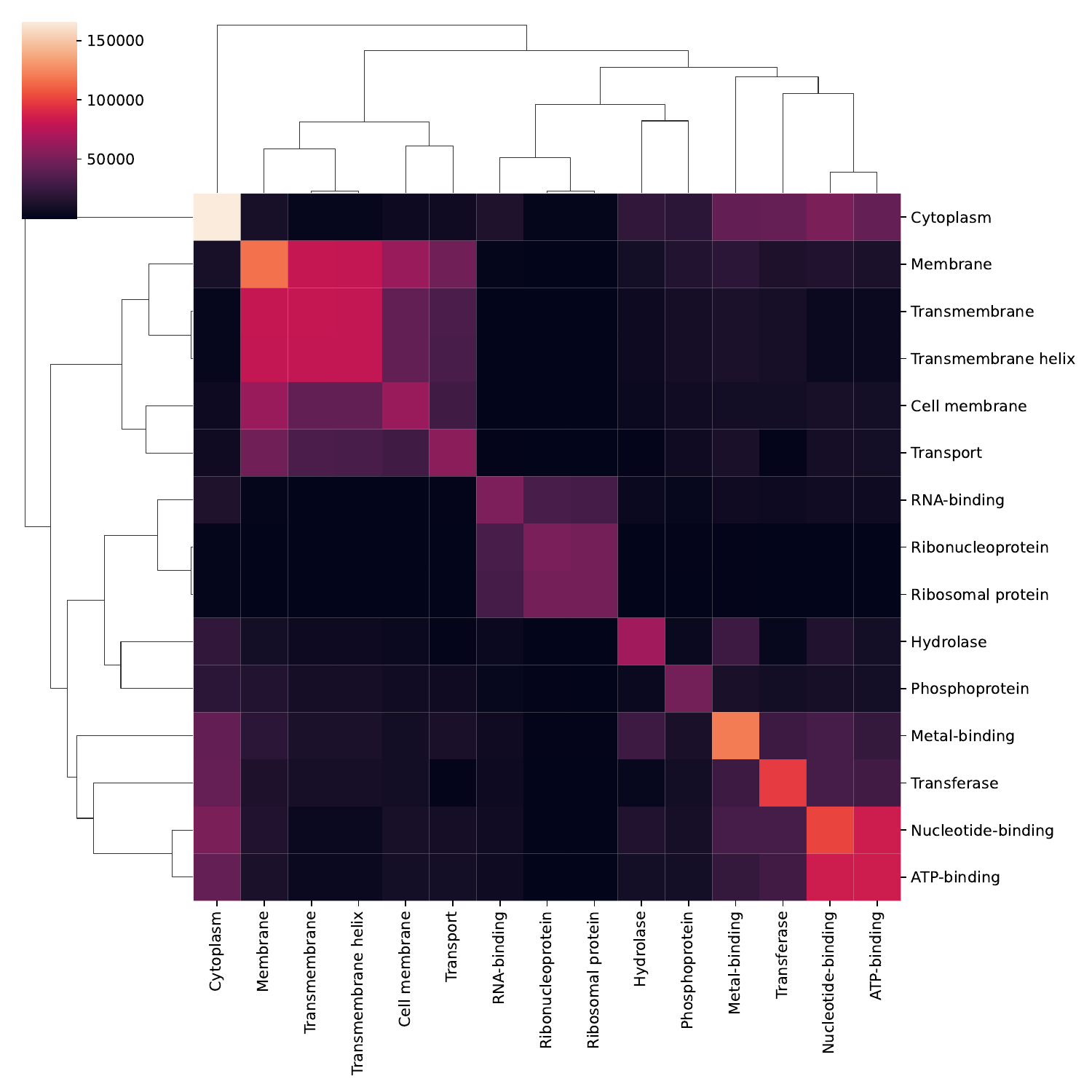}
    \caption{
    Swiss-Prot~\cite{bairoch1991swiss} protein keyword cooccurences.
    To construct the \textit{Cyto} and \textit{Binding} datasets, we search for popular but non-cooccuring keywords.
    }
    \label{fig:uniprot_keyword_coocurences}
\end{figure}

\subsection{fMRI experiment details}
\label{sec:fmri_supp}

This section gives more details on the fMRI experiment analyzed in \cref{sec:science}; for more scientific details see the original study~\citep{huth2016natural} and code (\href{https://github.com/HuthLab/speechmodeltutorial}{github.com/HuthLab/speechmodeltutorial}).
\cref{sec:science} analyzes data from one human subject in the original study,
as the subject listened to approximately two hours of narrative speech from the Moth Radio Hour, which consists of short autobiographical stories.
The subject underwent fMRI scanning as they listened, yielding an fMRI volume brain scan consisting of tens of thousands of voxels roughly every two seconds.

The individual voxel models described in \cref{sec:science} are each fit to 3,737 training points, each corresponding to a different time point (after accounting for various preprocessing steps, such as trimming the beginning and end of the sequence).
They are evaluated on 291 training volumes which come from a 10-minute story that was not seen during draining.

\begin{figure}[H]
    \centering
    \includegraphics[width=\textwidth]{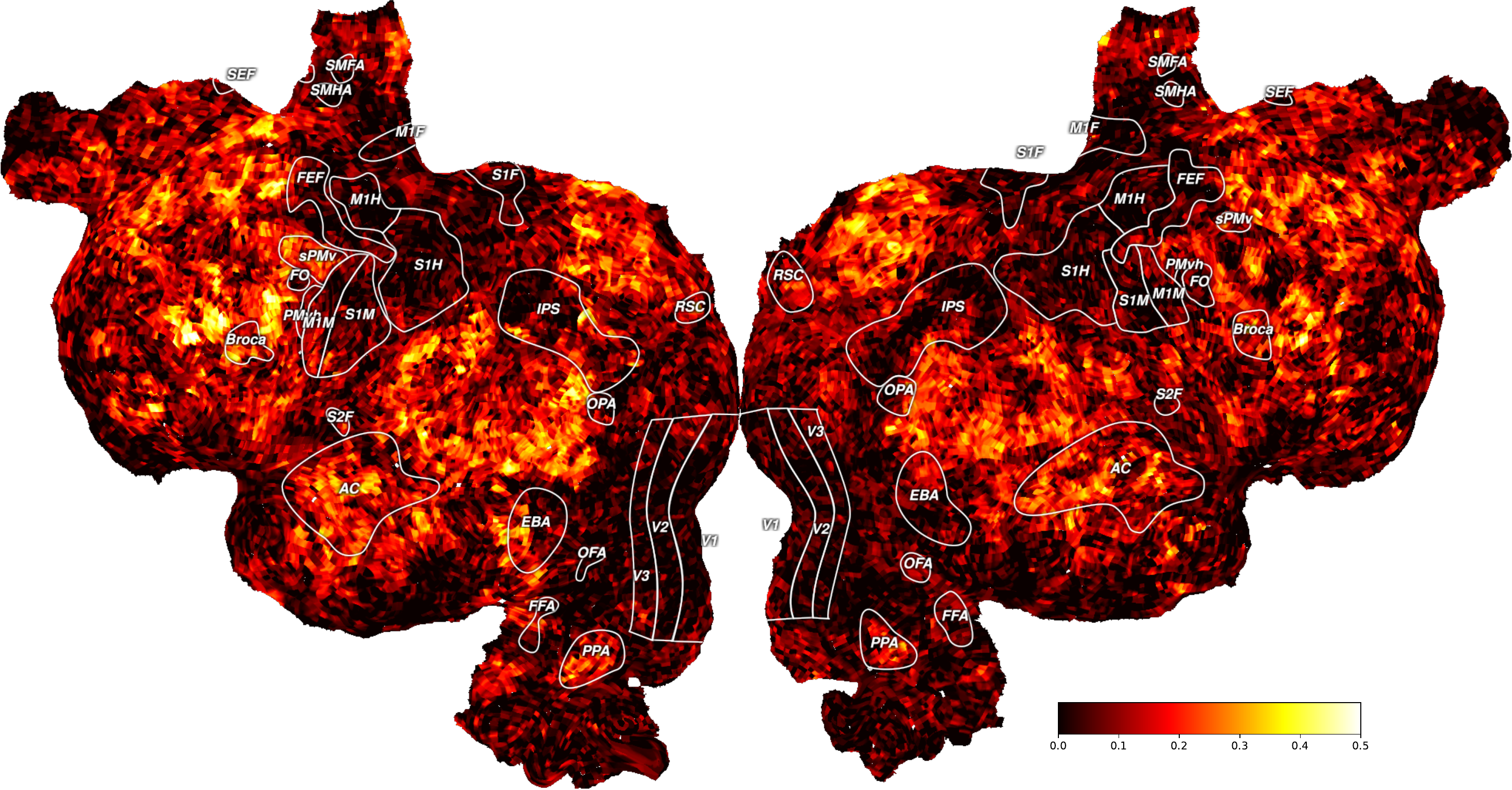}
    \caption{Generalization performance for individual-voxel models, measured by correlation between the prediction and the measured response.}
    \label{fig:flatmap_corr}
\end{figure}

\cref{fig:flatmap_corr} shows the generalization performance of the model for each voxel, measured by the correlation between the predicted response and the measured response.
Some regions are very poorly predicted (black), but many voxels can be predicted quite well (bright).

\begin{figure}[H]
    \centering
    \vspace{-25pt}
     \makebox[\textwidth][c]{\includegraphics[width=1.0\textwidth]{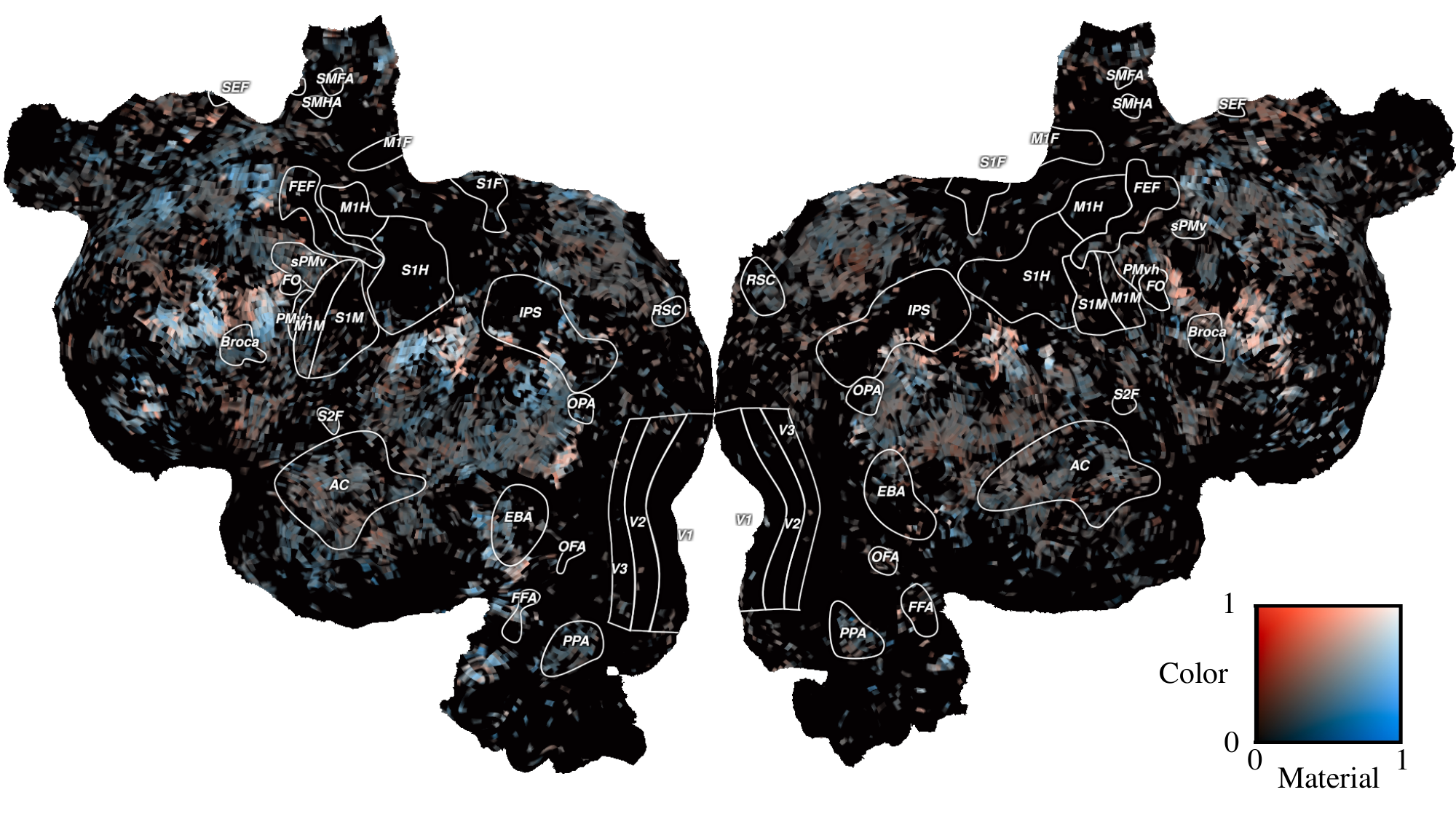}}%
     \vspace{-20pt}
    \caption{Representations of the \method-elicited concepts \textit{material} (blue) and \textit{color} (red) across the surface of the neocortex are spatially clustered and smooth.
    Left hemisphere corresponds to \cref{fig:flatmap}.
    % Each point shows the inferred voxel selectivity of the categories \textit{material} and \textit{color}, measured by an LLM.
    Only the top 10,000 best-predicted voxels are shown, remaining voxels are shown in black.
    Plotted with pycortex~\citep{gao2015pycortex}.}
    \label{fig:flatmap_supp}
\end{figure}

\begin{figure}[H]
    \centering
    \includegraphics[width=0.7\textwidth]{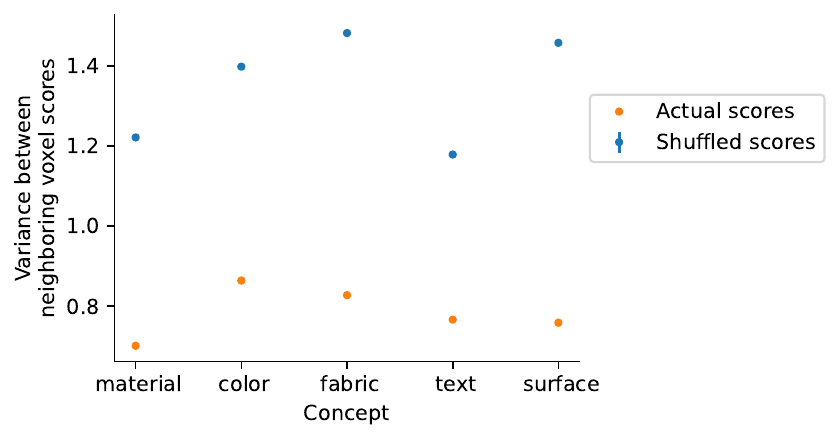}
    \caption{Concepts are spatially localized in the brain maps: the variance between neighboring voxels is considerably lower than would be expected from shuffling the voxel values.
    Note that we take care ot shuffle the map values only within the 10,000 top-predicted voxels, ignoring the poorly predicted voxels.
    Error bars (within the points) are standard errors of the mean.}
    \label{fig:concepts_clustered}
\end{figure}


\begin{thebibliography}{71}
\providecommand{\natexlab}[1]{#1}
\providecommand{\url}[1]{\texttt{#1}}
\expandafter\ifx\csname urlstyle\endcsname\relax
  \providecommand{\doi}[1]{doi: #1}\else
  \providecommand{\doi}{doi: \begingroup \urlstyle{rm}\Url}\fi

\bibitem[Agarwal et~al.(2022)Agarwal, Tan, Ronen, Singh, and
  Yu]{agarwal2022Hierarchical}
Agarwal, A., Tan, Y.~S., Ronen, O., Singh, C., and Yu, B.
\newblock Hierarchical shrinkage: improving the accuracy and interpretability
  of tree-based methods.
\newblock \emph{arXiv:2202.00858 [cs, stat]}, 2 2022.
\newblock URL \url{http://arxiv.org/abs/2202.00858}.
\newblock arXiv: 2202.00858.

\bibitem[Augusto \& Barbosa(2000)Augusto and Barbosa]{augusto2000symbolic}
Augusto, D.~A. and Barbosa, H.~J.
\newblock Symbolic regression via genetic programming.
\newblock In \emph{Proceedings. Vol. 1. Sixth Brazilian Symposium on Neural
  Networks}, pp.\  173--178. IEEE, 2000.

\bibitem[Bach et~al.(2022)Bach, Sanh, Yong, Webson, Raffel, Nayak, Sharma, Kim,
  Bari, Fevry, et~al.]{bach2022promptsource}
Bach, S.~H., Sanh, V., Yong, Z.-X., Webson, A., Raffel, C., Nayak, N.~V.,
  Sharma, A., Kim, T., Bari, M.~S., Fevry, T., et~al.
\newblock Promptsource: An integrated development environment and repository
  for natural language prompts.
\newblock \emph{arXiv preprint arXiv:2202.01279}, 2022.

\bibitem[Bairoch \& Boeckmann(1991)Bairoch and Boeckmann]{bairoch1991swiss}
Bairoch, A. and Boeckmann, B.
\newblock The swiss-prot protein sequence data bank.
\newblock \emph{Nucleic acids research}, 19\penalty0 (Suppl):\penalty0 2247,
  1991.

\bibitem[Beltagy et~al.(2019)Beltagy, Lo, and Cohan]{beltagy2019scibert}
Beltagy, I., Lo, K., and Cohan, A.
\newblock Scibert: A pretrained language model for scientific text.
\newblock \emph{arXiv preprint arXiv:1903.10676}, 2019.

\bibitem[Black et~al.(2021)Black, Leo, Wang, Leahy, and Biderman]{gpt_neo}
Black, S., Leo, G., Wang, P., Leahy, C., and Biderman, S.
\newblock {GPT-Neo: Large Scale Autoregressive Language Modeling with
  Mesh-Tensorflow}.
\newblock March 2021.
\newblock \doi{10.5281/zenodo.5297715}.
\newblock URL \url{https://doi.org/10.5281/zenodo.5297715}.
\newblock {If you use this software, please cite it using these metadata.}

\bibitem[Black et~al.(2022)Black, Biderman, Hallahan, Anthony, Gao, Golding,
  He, Leahy, McDonell, Phang, et~al.]{black2022gpt}
Black, S., Biderman, S., Hallahan, E., Anthony, Q., Gao, L., Golding, L., He,
  H., Leahy, C., McDonell, K., Phang, J., et~al.
\newblock Gpt-neox-20b: An open-source autoregressive language model.
\newblock \emph{arXiv preprint arXiv:2204.06745}, 2022.

\bibitem[Breiman et~al.(1984)Breiman, Friedman, Olshen, and
  Stone]{breiman1984classification}
Breiman, L., Friedman, J.~H., Olshen, R.~A., and Stone, C.~J.
\newblock \emph{Classification and Regression Trees}.
\newblock Wadsworth and Brooks, Monterey, CA, 1984.
\newblock URL
  \url{https://www.routledge.com/Classification-and-Regression-Trees/Breiman-Friedman-Stone-Olshen/p/book/9780412048418}.

\bibitem[Brown et~al.(2020)Brown, Mann, Ryder, Subbiah, Kaplan, Dhariwal,
  Neelakantan, Shyam, Sastry, Askell, et~al.]{brown2020language}
Brown, T., Mann, B., Ryder, N., Subbiah, M., Kaplan, J.~D., Dhariwal, P.,
  Neelakantan, A., Shyam, P., Sastry, G., Askell, A., et~al.
\newblock Language models are few-shot learners.
\newblock \emph{Advances in neural information processing systems},
  33:\penalty0 1877--1901, 2020.

\bibitem[Camburu et~al.(2018)Camburu, Rockt{\"a}schel, Lukasiewicz, and
  Blunsom]{camburu2018snli}
Camburu, O.-M., Rockt{\"a}schel, T., Lukasiewicz, T., and Blunsom, P.
\newblock e-snli: Natural language inference with natural language
  explanations.
\newblock \emph{Advances in Neural Information Processing Systems}, 31, 2018.

\bibitem[Chung et~al.(2022)Chung, Hou, Longpre, Zoph, Tay, Fedus, Li, Wang,
  Dehghani, Brahma, Webson, Gu, Dai, Suzgun, Chen, Chowdhery, Narang, Mishra,
  Yu, Zhao, Huang, Dai, Yu, Petrov, Chi, Dean, Devlin, Roberts, Zhou, Le, and
  Wei]{flan}
Chung, H.~W., Hou, L., Longpre, S., Zoph, B., Tay, Y., Fedus, W., Li, E., Wang,
  X., Dehghani, M., Brahma, S., Webson, A., Gu, S.~S., Dai, Z., Suzgun, M.,
  Chen, X., Chowdhery, A., Narang, S., Mishra, G., Yu, A., Zhao, V., Huang, Y.,
  Dai, A., Yu, H., Petrov, S., Chi, E.~H., Dean, J., Devlin, J., Roberts, A.,
  Zhou, D., Le, Q.~V., and Wei, J.
\newblock Scaling instruction-finetuned language models, 2022.
\newblock URL \url{https://arxiv.org/abs/2210.11416}.

\bibitem[Conneau et~al.(2018)Conneau, Kruszewski, Lample, Barrault, and
  Baroni]{conneau2018you}
Conneau, A., Kruszewski, G., Lample, G., Barrault, L., and Baroni, M.
\newblock What you can cram into a single vector: Probing sentence embeddings
  for linguistic properties.
\newblock \emph{arXiv preprint arXiv:1805.01070}, 2018.

\bibitem[Consortium(2015)]{uniprot2015uniprot}
Consortium, U.
\newblock Uniprot: a hub for protein information.
\newblock \emph{Nucleic acids research}, 43\penalty0 (D1):\penalty0 D204--D212,
  2015.

\bibitem[Deng et~al.(2022)Deng, Wang, Hsieh, Wang, Guo, Shu, Song, Xing, and
  Hu]{deng2022rlprompt}
Deng, M., Wang, J., Hsieh, C.-P., Wang, Y., Guo, H., Shu, T., Song, M., Xing,
  E.~P., and Hu, Z.
\newblock Rlprompt: Optimizing discrete text prompts with reinforcement
  learning.
\newblock \emph{arXiv preprint arXiv:2205.12548}, 2022.

\bibitem[Devlin et~al.(2018)Devlin, Chang, Lee, and Toutanova]{devlin2018bert}
Devlin, J., Chang, M.-W., Lee, K., and Toutanova, K.
\newblock Bert: Pre-training of deep bidirectional transformers for language
  understanding.
\newblock \emph{arXiv preprint arXiv:1810.04805}, 2018.

\bibitem[Gao et~al.(2015)Gao, Huth, Lescroart, and Gallant]{gao2015pycortex}
Gao, J.~S., Huth, A.~G., Lescroart, M.~D., and Gallant, J.~L.
\newblock Pycortex: an interactive surface visualizer for fmri.
\newblock \emph{Frontiers in neuroinformatics}, pp.\ ~23, 2015.

\bibitem[Gulwani et~al.(2017)Gulwani, Polozov, Singh,
  et~al.]{gulwani2017program}
Gulwani, S., Polozov, O., Singh, R., et~al.
\newblock Program synthesis.
\newblock \emph{Foundations and Trends{\textregistered} in Programming
  Languages}, 4\penalty0 (1-2):\penalty0 1--119, 2017.

\bibitem[Ha et~al.(2021)Ha, Singh, Lanusse, Upadhyayula, and
  Yu]{ha2021adaptive}
Ha, W., Singh, C., Lanusse, F., Upadhyayula, S., and Yu, B.
\newblock Adaptive wavelet distillation from neural networks through
  interpretations.
\newblock \emph{Advances in Neural Information Processing Systems}, 34, 2021.

\bibitem[Han et~al.(2021)Han, Zhao, Ding, Liu, and Sun]{han2021ptr}
Han, X., Zhao, W., Ding, N., Liu, Z., and Sun, M.
\newblock Ptr: Prompt tuning with rules for text classification.
\newblock \emph{arXiv preprint arXiv:2105.11259}, 2021.

\bibitem[Hand(2007)]{hand2007principles}
Hand, D.~J.
\newblock Principles of data mining.
\newblock \emph{Drug safety}, 30\penalty0 (7):\penalty0 621--622, 2007.

\bibitem[Hendricks et~al.(2016)Hendricks, Akata, Rohrbach, Donahue, Schiele,
  and Darrell]{hendricks2016generating}
Hendricks, L.~A., Akata, Z., Rohrbach, M., Donahue, J., Schiele, B., and
  Darrell, T.
\newblock Generating visual explanations.
\newblock In \emph{European conference on computer vision}, pp.\  3--19.
  Springer, 2016.

\bibitem[Holtzman et~al.(2019)Holtzman, Buys, Du, Forbes, and
  Choi]{holtzman2019curious}
Holtzman, A., Buys, J., Du, L., Forbes, M., and Choi, Y.
\newblock The curious case of neural text degeneration.
\newblock \emph{arXiv preprint arXiv:1904.09751}, 2019.

\bibitem[Honovich et~al.(2022)Honovich, Shaham, Bowman, and
  Levy]{honovich2022instruction}
Honovich, O., Shaham, U., Bowman, S.~R., and Levy, O.
\newblock Instruction induction: From few examples to natural language task
  descriptions.
\newblock \emph{arXiv preprint arXiv:2205.10782}, 2022.

\bibitem[Hu et~al.(2021)Hu, Ding, Wang, Liu, Li, and Sun]{hu2021knowledgeable}
Hu, S., Ding, N., Wang, H., Liu, Z., Li, J., and Sun, M.
\newblock Knowledgeable prompt-tuning: Incorporating knowledge into prompt
  verbalizer for text classification.
\newblock \emph{arXiv preprint arXiv:2108.02035}, 2021.

\bibitem[Huth et~al.(2016)Huth, De~Heer, Griffiths, Theunissen, and
  Gallant]{huth2016natural}
Huth, A.~G., De~Heer, W.~A., Griffiths, T.~L., Theunissen, F.~E., and Gallant,
  J.~L.
\newblock Natural speech reveals the semantic maps that tile human cerebral
  cortex.
\newblock \emph{Nature}, 532\penalty0 (7600):\penalty0 453--458, 2016.

\bibitem[Kry{\'s}ci{\'n}ski et~al.(2019)Kry{\'s}ci{\'n}ski, Keskar, McCann,
  Xiong, and Socher]{kryscinski2019neural}
Kry{\'s}ci{\'n}ski, W., Keskar, N.~S., McCann, B., Xiong, C., and Socher, R.
\newblock Neural text summarization: A critical evaluation.
\newblock \emph{arXiv preprint arXiv:1908.08960}, 2019.

\bibitem[Lewkowycz et~al.(2022)Lewkowycz, Andreassen, Dohan, Dyer, Michalewski,
  Ramasesh, Slone, Anil, Schlag, Gutman-Solo, et~al.]{lewkowycz2022solving}
Lewkowycz, A., Andreassen, A., Dohan, D., Dyer, E., Michalewski, H., Ramasesh,
  V., Slone, A., Anil, C., Schlag, I., Gutman-Solo, T., et~al.
\newblock Solving quantitative reasoning problems with language models.
\newblock \emph{arXiv preprint arXiv:2206.14858}, 2022.

\bibitem[Li \& Liang(2021)Li and Liang]{li2021prefix}
Li, X.~L. and Liang, P.
\newblock Prefix-tuning: Optimizing continuous prompts for generation.
\newblock \emph{arXiv preprint arXiv:2101.00190}, 2021.

\bibitem[Liu \& Avci(2019)Liu and Avci]{liu2019incorporating}
Liu, F. and Avci, B.
\newblock Incorporating priors with feature attribution on text classification.
\newblock \emph{arXiv preprint arXiv:1906.08286}, 2019.

\bibitem[Liu et~al.(2021{\natexlab{a}})Liu, Yuan, Fu, Jiang, Hayashi, and
  Neubig]{liu2021pre}
Liu, P., Yuan, W., Fu, J., Jiang, Z., Hayashi, H., and Neubig, G.
\newblock Pre-train, prompt, and predict: A systematic survey of prompting
  methods in natural language processing.
\newblock \emph{arXiv preprint arXiv:2107.13586}, 2021{\natexlab{a}}.

\bibitem[Liu et~al.(2018)Liu, Wang, Sha, Chang, and Sui]{liu2018table}
Liu, T., Wang, K., Sha, L., Chang, B., and Sui, Z.
\newblock Table-to-text generation by structure-aware seq2seq learning.
\newblock In \emph{Thirty-Second AAAI Conference on Artificial Intelligence},
  2018.

\bibitem[Liu et~al.(2021{\natexlab{b}})Liu, Zheng, Du, Ding, Qian, Yang, and
  Tang]{liu2021gpt}
Liu, X., Zheng, Y., Du, Z., Ding, M., Qian, Y., Yang, Z., and Tang, J.
\newblock Gpt understands, too.
\newblock \emph{arXiv preprint arXiv:2103.10385}, 2021{\natexlab{b}}.

\bibitem[Logan~IV et~al.(2022)Logan~IV, Balazevic, Wallace, Petroni, Singh, and
  Riedel]{logan-iv-etal-2022-cutting}
Logan~IV, R., Balazevic, I., Wallace, E., Petroni, F., Singh, S., and Riedel,
  S.
\newblock Cutting down on prompts and parameters: Simple few-shot learning with
  language models.
\newblock In \emph{Findings of the Association for Computational Linguistics:
  ACL 2022}, pp.\  2824--2835, Dublin, Ireland, May 2022. Association for
  Computational Linguistics.
\newblock \doi{10.18653/v1/2022.findings-acl.222}.
\newblock URL \url{https://aclanthology.org/2022.findings-acl.222}.

\bibitem[Lu et~al.(2022)Lu, Bartolo, Moore, Riedel, and
  Stenetorp]{lu-etal-2022-fantastically}
Lu, Y., Bartolo, M., Moore, A., Riedel, S., and Stenetorp, P.
\newblock Fantastically ordered prompts and where to find them: Overcoming
  few-shot prompt order sensitivity.
\newblock In \emph{Proceedings of the 60th Annual Meeting of the Association
  for Computational Linguistics (Volume 1: Long Papers)}, pp.\  8086--8098,
  Dublin, Ireland, May 2022. Association for Computational Linguistics.
\newblock \doi{10.18653/v1/2022.acl-long.556}.
\newblock URL \url{https://aclanthology.org/2022.acl-long.556}.

\bibitem[Lundberg et~al.(2019)Lundberg, Erion, Chen, DeGrave, Prutkin, Nair,
  Katz, Himmelfarb, Bansal, and Lee]{lundberg2019explainable}
Lundberg, S.~M., Erion, G., Chen, H., DeGrave, A., Prutkin, J.~M., Nair, B.,
  Katz, R., Himmelfarb, J., Bansal, N., and Lee, S.-I.
\newblock Explainable ai for trees: From local explanations to global
  understanding.
\newblock \emph{arXiv preprint arXiv:1905.04610}, 2019.

\bibitem[Malo et~al.(2014)Malo, Sinha, Korhonen, Wallenius, and
  Takala]{Malo2014GoodDO}
Malo, P., Sinha, A., Korhonen, P., Wallenius, J., and Takala, P.
\newblock Good debt or bad debt: Detecting semantic orientations in economic
  texts.
\newblock \emph{Journal of the Association for Information Science and
  Technology}, 65, 2014.

\bibitem[Manna \& Waldinger(1980)Manna and Waldinger]{manna1980deductive}
Manna, Z. and Waldinger, R.
\newblock A deductive approach to program synthesis.
\newblock \emph{ACM Transactions on Programming Languages and Systems
  (TOPLAS)}, 2\penalty0 (1):\penalty0 90--121, 1980.

\bibitem[Meng et~al.(2022)Meng, Bau, Andonian, and Belinkov]{meng2022locating}
Meng, K., Bau, D., Andonian, A., and Belinkov, Y.
\newblock Locating and editing factual knowledge in gpt.
\newblock \emph{arXiv preprint arXiv:2202.05262}, 2022.

\bibitem[Olah et~al.(2018)Olah, Satyanarayan, Johnson, Carter, Schubert, Ye,
  and Mordvintsev]{olah2018building}
Olah, C., Satyanarayan, A., Johnson, I., Carter, S., Schubert, L., Ye, K., and
  Mordvintsev, A.
\newblock The building blocks of interpretability.
\newblock \emph{Distill}, 3\penalty0 (3):\penalty0 e10, 2018.

\bibitem[Pang \& Lee(2005)Pang and Lee]{PangLee2005}
Pang, B. and Lee, L.
\newblock Seeing stars: Exploiting class relationships for sentiment
  categorization with respect to rating scales.
\newblock In \emph{Proceedings of the ACL}, 2005.

\bibitem[Petroni et~al.(2019)Petroni, Rockt{\"a}schel, Lewis, Bakhtin, Wu,
  Miller, and Riedel]{petroni2019language}
Petroni, F., Rockt{\"a}schel, T., Lewis, P., Bakhtin, A., Wu, Y., Miller,
  A.~H., and Riedel, S.
\newblock Language models as knowledge bases?
\newblock \emph{arXiv preprint arXiv:1909.01066}, 2019.

\bibitem[Radford et~al.(2019)Radford, Wu, Child, Luan, Amodei, Sutskever,
  et~al.]{radford2019language}
Radford, A., Wu, J., Child, R., Luan, D., Amodei, D., Sutskever, I., et~al.
\newblock Language models are unsupervised multitask learners.
\newblock \emph{OpenAI blog}, 1\penalty0 (8):\penalty0 9, 2019.

\bibitem[Radford et~al.(2021)Radford, Kim, Hallacy, Ramesh, Goh, Agarwal,
  Sastry, Askell, Mishkin, Clark, et~al.]{radford2021learning}
Radford, A., Kim, J.~W., Hallacy, C., Ramesh, A., Goh, G., Agarwal, S., Sastry,
  G., Askell, A., Mishkin, P., Clark, J., et~al.
\newblock Learning transferable visual models from natural language
  supervision.
\newblock In \emph{International Conference on Machine Learning}, pp.\
  8748--8763. PMLR, 2021.

\bibitem[Raffel et~al.(2020)Raffel, Shazeer, Roberts, Lee, Narang, Matena,
  Zhou, Li, Liu, et~al.]{raffel2020exploring}
Raffel, C., Shazeer, N., Roberts, A., Lee, K., Narang, S., Matena, M., Zhou,
  Y., Li, W., Liu, P.~J., et~al.
\newblock Exploring the limits of transfer learning with a unified text-to-text
  transformer.
\newblock \emph{J. Mach. Learn. Res.}, 21\penalty0 (140):\penalty0 1--67, 2020.

\bibitem[Ribeiro et~al.(2016)Ribeiro, Singh, and Guestrin]{ribeiro2016should}
Ribeiro, M.~T., Singh, S., and Guestrin, C.
\newblock Why should i trust you?: Explaining the predictions of any
  classifier.
\newblock In \emph{Proceedings of the 22nd ACM SIGKDD International Conference
  on Knowledge Discovery and Data Mining}, pp.\  1135--1144. ACM, 2016.

\bibitem[Richard et~al.(2020)Richard, Huang, Waidyanatha, Shinn, Collins,
  Thillainadarajah, Grulke, Williams, Lougee, Judson, et~al.]{richard2020tox21}
Richard, A.~M., Huang, R., Waidyanatha, S., Shinn, P., Collins, B.~J.,
  Thillainadarajah, I., Grulke, C.~M., Williams, A.~J., Lougee, R.~R., Judson,
  R.~S., et~al.
\newblock The tox21 10k compound library: collaborative chemistry advancing
  toxicology.
\newblock \emph{Chemical Research in Toxicology}, 34\penalty0 (2):\penalty0
  189--216, 2020.

\bibitem[Sadat \& Caragea(2022)Sadat and Caragea]{sadat2022scinli}
Sadat, M. and Caragea, C.
\newblock Scinli: A corpus for natural language inference on scientific text.
\newblock \emph{arXiv preprint arXiv:2203.06728}, 2022.

\bibitem[Sanh et~al.(2021)Sanh, Webson, Raffel, Bach, Sutawika, Alyafeai,
  Chaffin, Stiegler, Scao, Raja, et~al.]{sanh2021multitask}
Sanh, V., Webson, A., Raffel, C., Bach, S.~H., Sutawika, L., Alyafeai, Z.,
  Chaffin, A., Stiegler, A., Scao, T.~L., Raja, A., et~al.
\newblock Multitask prompted training enables zero-shot task generalization.
\newblock \emph{arXiv preprint arXiv:2110.08207}, 2021.

\bibitem[Schmidt \& Lipson(2009)Schmidt and Lipson]{schmidt2009distilling}
Schmidt, M. and Lipson, H.
\newblock Distilling free-form natural laws from experimental data.
\newblock \emph{science}, 324\penalty0 (5923):\penalty0 81--85, 2009.

\bibitem[Schrimpf et~al.(2021)Schrimpf, Blank, Tuckute, Kauf, Hosseini,
  Kanwisher, Tenenbaum, and Fedorenko]{schrimpf2021neural}
Schrimpf, M., Blank, I.~A., Tuckute, G., Kauf, C., Hosseini, E.~A., Kanwisher,
  N., Tenenbaum, J.~B., and Fedorenko, E.
\newblock The neural architecture of language: Integrative modeling converges
  on predictive processing.
\newblock \emph{Proceedings of the National Academy of Sciences}, 118\penalty0
  (45):\penalty0 e2105646118, 2021.

\bibitem[Sha et~al.(2021)Sha, Camburu, and Lukasiewicz]{sha2021learning}
Sha, L., Camburu, O.-M., and Lukasiewicz, T.
\newblock Learning from the best: Rationalizing predictions by adversarial
  information calibration.
\newblock In \emph{AAAI}, pp.\  13771--13779, 2021.

\bibitem[Shin et~al.(2020)Shin, Razeghi, Logan~IV, Wallace, and
  Singh]{shin2020autoprompt}
Shin, T., Razeghi, Y., Logan~IV, R.~L., Wallace, E., and Singh, S.
\newblock Autoprompt: Eliciting knowledge from language models with
  automatically generated prompts.
\newblock \emph{arXiv preprint arXiv:2010.15980}, 2020.

\bibitem[Singh \& Gao(2022)Singh and Gao]{singh2022embgam}
Singh, C. and Gao, J.
\newblock Emb-gam: an interpretable and efficient predictor using pre-trained
  language models.
\newblock \emph{arXiv preprint arXiv:2209.11799}, 2022.
\newblock \doi{10.48550/arxiv.2209.11799}.
\newblock URL \url{https://arxiv.org/abs/2209.11799}.

\bibitem[Singh et~al.(2019)Singh, Murdoch, and Yu]{singh2019Hierarchical}
Singh, C., Murdoch, W.~J., and Yu, B.
\newblock Hierarchical interpretations for neural network predictions.
\newblock \emph{International Conference on Learning Representations}, pp.\
  ~26, 2019.
\newblock URL \url{https://openreview.net/forum?id=SkEqro0ctQ}.

\bibitem[Singh et~al.(2021)Singh, Nasseri, Tan, Tang, and Yu]{singh2021imodels}
Singh, C., Nasseri, K., Tan, Y.~S., Tang, T., and Yu, B.
\newblock imodels: a python package for fitting interpretable models.
\newblock \emph{Journal of Open Source Software}, 6\penalty0 (61):\penalty0
  3192, 2021.
\newblock \doi{10.21105/joss.03192}.
\newblock URL \url{https://doi.org/10.21105/joss.03192}.

\bibitem[Socher et~al.(2013)Socher, Perelygin, Wu, Chuang, Manning, Ng, and
  Potts]{socher2013recursive}
Socher, R., Perelygin, A., Wu, J., Chuang, J., Manning, C.~D., Ng, A., and
  Potts, C.
\newblock Recursive deep models for semantic compositionality over a sentiment
  treebank.
\newblock In \emph{Proceedings of the 2013 conference on empirical methods in
  natural language processing}, pp.\  1631--1642, 2013.

\bibitem[Strobelt et~al.(2022)Strobelt, Webson, Sanh, Hoover, Beyer, Pfister,
  and Rush]{strobelt2022interactive}
Strobelt, H., Webson, A., Sanh, V., Hoover, B., Beyer, J., Pfister, H., and
  Rush, A.~M.
\newblock Interactive and visual prompt engineering for ad-hoc task adaptation
  with large language models.
\newblock \emph{arXiv preprint arXiv:2208.07852}, 2022.

\bibitem[Tan et~al.(2018)Tan, Caruana, Hooker, and Lou]{tan2018distill}
Tan, S., Caruana, R., Hooker, G., and Lou, Y.
\newblock Distill-and-compare: Auditing black-box models using transparent
  model distillation.
\newblock In \emph{Proceedings of the 2018 AAAI/ACM Conference on AI, Ethics,
  and Society}, pp.\  303--310, 2018.

\bibitem[Tan et~al.(2022)Tan, Singh, Nasseri, Agarwal, and Yu]{tan2022Fast}
Tan, Y.~S., Singh, C., Nasseri, K., Agarwal, A., and Yu, B.
\newblock Fast interpretable greedy-tree sums (figs).
\newblock \emph{arXiv:2201.11931 [cs, stat]}, 1 2022.
\newblock URL \url{http://arxiv.org/abs/2201.11931}.
\newblock arXiv: 2201.11931.

\bibitem[Taylor et~al.(2022)Taylor, Kardas, Cucurull, Scialom, Hartshorn,
  Saravia, Poulton, Kerkez, and Stojnic]{taylor2022galactica}
Taylor, R., Kardas, M., Cucurull, G., Scialom, T., Hartshorn, A., Saravia, E.,
  Poulton, A., Kerkez, V., and Stojnic, R.
\newblock Galactica: A large language model for science.
\newblock \emph{arXiv preprint arXiv:2211.09085}, 2022.

\bibitem[Tsang et~al.(2017)Tsang, Cheng, and Liu]{tsang2017detecting}
Tsang, M., Cheng, D., and Liu, Y.
\newblock Detecting statistical interactions from neural network weights.
\newblock \emph{arXiv preprint arXiv:1705.04977}, 2017.

\bibitem[Wallace et~al.(2019)Wallace, Feng, Kandpal, Gardner, and
  Singh]{wallace2019universal}
Wallace, E., Feng, S., Kandpal, N., Gardner, M., and Singh, S.
\newblock Universal adversarial triggers for attacking and analyzing nlp.
\newblock \emph{arXiv preprint arXiv:1908.07125}, 2019.

\bibitem[Wang \& Komatsuzaki(2021)Wang and Komatsuzaki]{gpt_j}
Wang, B. and Komatsuzaki, A.
\newblock {GPT-J-6B: A 6 Billion Parameter Autoregressive Language Model}.
\newblock \url{https://github.com/kingoflolz/mesh-transformer-jax}, May 2021.

\bibitem[Wang et~al.(2021)Wang, Xu, Tong, Roberts, and Liu]{wang2021inferbert}
Wang, X., Xu, X., Tong, W., Roberts, R., and Liu, Z.
\newblock Inferbert: a transformer-based causal inference framework for
  enhancing pharmacovigilance.
\newblock \emph{Frontiers in Artificial Intelligence}, 4:\penalty0 659622,
  2021.

\bibitem[Wang et~al.(2022)Wang, Mishra, Alipoormolabashi, Kordi,
  et~al.]{naturalinstructionsv2}
Wang, Y., Mishra, S., Alipoormolabashi, P., Kordi, Y., et~al.
\newblock Benchmarking generalization via in-context instructions on 1,600+
  language tasks.
\newblock \emph{arXiv}, 2022.

\bibitem[Webson \& Pavlick(2021)Webson and Pavlick]{webson2021prompt}
Webson, A. and Pavlick, E.
\newblock Do prompt-based models really understand the meaning of their
  prompts?
\newblock \emph{arXiv preprint arXiv:2109.01247}, 2021.

\bibitem[Zaidan \& Eisner(2008)Zaidan and Eisner]{zaidan2008modeling}
Zaidan, O. and Eisner, J.
\newblock Modeling annotators: A generative approach to learning from annotator
  rationales.
\newblock In \emph{Proceedings of the 2008 conference on Empirical methods in
  natural language processing}, pp.\  31--40, 2008.

\bibitem[Zhang et~al.(2022)Zhang, Roller, Goyal, Artetxe, Chen, Chen, Dewan,
  Diab, Li, Lin, et~al.]{zhang2022opt}
Zhang, S., Roller, S., Goyal, N., Artetxe, M., Chen, M., Chen, S., Dewan, C.,
  Diab, M., Li, X., Lin, X.~V., et~al.
\newblock Opt: Open pre-trained transformer language models.
\newblock \emph{arXiv preprint arXiv:2205.01068}, 2022.

\bibitem[Zhong et~al.(2021)Zhong, Lee, Zhang, and Klein]{zhong2021adapting}
Zhong, R., Lee, K., Zhang, Z., and Klein, D.
\newblock Adapting language models for zero-shot learning by meta-tuning on
  dataset and prompt collections.
\newblock \emph{arXiv preprint arXiv:2104.04670}, 2021.

\bibitem[Zhong et~al.(2022)Zhong, Snell, Klein, and
  Steinhardt]{zhong2022describing}
Zhong, R., Snell, C., Klein, D., and Steinhardt, J.
\newblock Describing differences between text distributions with natural
  language.
\newblock In \emph{International Conference on Machine Learning}, pp.\
  27099--27116. PMLR, 2022.

\bibitem[Zhou et~al.(2022)Zhou, Muresanu, Han, Paster, Pitis, Chan, and
  Ba]{zhou2022large}
Zhou, Y., Muresanu, A.~I., Han, Z., Paster, K., Pitis, S., Chan, H., and Ba, J.
\newblock Large language models are human-level prompt engineers.
\newblock \emph{arXiv preprint arXiv:2211.01910}, 2022.

\end{thebibliography}
\end{document}